%% file: main.tex
\documentclass[ba,preprint]{imsart}

\usepackage{amsthm}
\usepackage{amsmath}
\usepackage{amssymb}
\usepackage{bm}
\usepackage{amsfonts}
\usepackage[ruled,vlined]{algorithm2e}
\usepackage{natbib}
\usepackage[colorlinks,citecolor=blue,urlcolor=blue,filecolor=blue,backref=page]{hyperref}
\usepackage{graphicx}
\usepackage{booktabs}
\usepackage{url}
\usepackage{placeins}
\usepackage{subcaption}
\startlocaldefs
\numberwithin{equation}{section}
\theoremstyle{plain}

\endlocaldefs

\theoremstyle{plain}
\newtheorem{theorem}{Theorem}[section]

\newtheorem{proposition}[theorem]{Proposition}

\theoremstyle{definition}

\begin{document}

\begin{frontmatter}
\pdftitle{Approximate Reverse-KL Transports for Reversible Jump MCMC}
\pdfauthor{Pingping Yin and Xiyun Jiao}
\pdfsubject{Research preprint}
\pdfkeywords{reversible jump MCMC, transport maps, normalizing flows, variational inference, trans-dimensional Bayesian computation}
\title{Approximate Reverse-KL Transports for Reversible Jump MCMC}
\runtitle{Approximate Reverse-KL Transports for RJMCMC}

\begin{aug}
\author{\fnms{Pingping} \snm{Yin}\thanksref{addr1}}
\and
\author{\fnms{Xiyun} \snm{Jiao}\thanksref{addr1,t1}
\ead[label=e1]{jiaoxy@sustech.edu.cn}}

\runauthor{P. Yin and X. Jiao}

\address[addr1]{Department of Statistics and Data Science, Southern University of Science and Technology}

\thankstext{t1}{Corresponding author
    \printead{e1}
}

\end{aug}

\begin{abstract}
We develop transport proposals for reversible jump Markov chain Monte Carlo (RJMCMC) using reverse-Kullback--Leibler variational inference with normalizing flows. Unlike sample-trained transport reversible-jump proposals, our map-fitting objectives use evaluations of unnormalized model-specific posterior densities and draws from tractable reference distributions, rather than pilot posterior samples. The fitted flows are used only to construct Metropolis--Hastings proposals; consequently, the resulting sampler targets the exact trans-dimensional posterior even when the variational transports are imperfect. A central contribution is a trans-dimensional variational-inference construction for conditional transport proposals. Model-specific parameters are embedded in a saturated space, and a conditional normalizing flow learns the corresponding transport maps jointly across model dimensions. We use RealNVP coupling flows, unlike autoregressive flows, because their coordinatewise map and inverse evaluations are parallelizable, which is well suited to repeated RJMCMC proposal evaluations. We derive a residual-weight representation of the between-model acceptance probability. It clarifies that reverse-KL under-coverage affects finite-time proposal quality and mixing, but not the invariant posterior distribution after Metropolis--Hastings correction. Experiments on a controlled sinh--arcsinh example, Bayesian factor analysis, robust-regression variable selection, coal-mining change-point inference, and phylogenetic substitution-model selection for two sequence alignments over 203 candidate models show that the proposed target-trained transports can substantially improve between-model mobility and posterior model-probability estimation. The trans-dimensional conditional method provides an effective joint alternative to particle-trained conditional transports, particularly in the larger model spaces considered.
\end{abstract}

\end{frontmatter}

\section{Introduction}

\label{sec:intro}
Many inference problems have a trans-model nature, such as variable selection~\citep{BS_fan2026reversible}, phylogenetic tree search~\citep{J_NSR_jiao2021multispecies}, geoscientific inversion~\citep{J_PTRSA_sambridge2013transdimensional}, etc. Denote the data by $\mathbf{y}$ and suppose there are $K$ ($\ge 2$) candidate models for $\mathbf{y}$. Each model $k$ has a vector of $d_k$ parameters $\boldsymbol{\theta}_k \in \boldsymbol{\Theta} _k\subset \mathbb{R}^{d_k}$, where $d_k$ may vary with $k$. Then the unknown parameter of the trans-model inference problem is $(k,\boldsymbol{\theta}_k)$, with the state space $\boldsymbol{\Omega}=\bigcup_{k=1}^K\left ( \left \{  k\right \} \times \boldsymbol{\Theta} _k \right)$, i.e., the union of the subspaces for the $K$ models. The joint posterior distribution of $(k,\boldsymbol{\theta}_k)$ is
\begin{equation}
\pi(k, \boldsymbol{\theta}_k \mid \mathbf{y}) \propto \pi(k) \, \pi(\boldsymbol{\theta}_k \mid k) \, \pi(\mathbf{y} \mid \boldsymbol{\theta}_k, k),
\label{eq:trans_model_posterior}
\end{equation}
where $\pi(k)$ is the prior of model index $k$, $\pi(\boldsymbol{\theta}_k \mid k)$ and $\pi(\mathbf{y} \mid \boldsymbol{\theta}_k, k)$ are the prior and likelihood under model $k$.

Reversible jump Markov Chain Monte Carlo \citep[RJMCMC,][]{J_BIOMET_green1995reversible}, a generalization of Metropolis-Hastings, is designed to sample from trans-model posteriors as eq.~\eqref{eq:trans_model_posterior}. It is notoriously difficult for RJMCMC to achieve reasonable acceptance rate and mixing efficiency when moving between models. Strategies to design efficient trans-model proposals include constructing multivariate Gaussian proposals analogous to random walks with informative means and variances \citep{BS_green2003transdimensional}, estimating conditional marginal densities \citep{J_SAC_fan2009automating}, using KD-tree approximation of the target densities for auxiliary-variable draws \citep{J_RSOS_farr2015efficient}, locally-informed proposals for model index \citep{J_EJS_gagnon2021informed}, etc. Other approaches to encouraging trans-model jumps involve multi-step proposals \citep{J_BIOMET_green2001delayed,J_SPL_al-awadhi2004improving} and saturated-space methods~\citep{J_JRSSSB_brooks2003efficient}. However, none of these methods are general enough to be widely applied in practice.

\citet{C_AISTATS_davies2023transport} extended the idea of using measure transports to accelerate MCMC in recent work~\citep{J_JUQ_parno2018transport,  arXiv_hoffman2019neutralizing, C_ICMLW_gabrie2021efficient} into trans-model scenarios, and successfully developed a general method for constructing efficient trans-dimensional proposals, named by \emph{transport reversible jump} (TRJ). To propose from model $k$ to $k'$, TRJ first (approximately) transports $\boldsymbol{\theta}_k$ from the conditional posterior $\pi_k=\pi(\cdot \mid k, \mathbf{y})$ to an auxiliary variable $\boldsymbol{z}_k$ from some easy-to-sample \emph{reference} distribution, proposes in the reference space to get a new candidate $\boldsymbol{z}_{k'}$ with dimension $d_{k'}$, and then transports $\boldsymbol{z}_{k'}$ to $\boldsymbol{\theta}_{k'}$, which is approximately $\pi_{k'}$-distributed. \citet{C_AISTATS_davies2023transport} proved that if the transports are exact, the resulting algorithm has the same acceptance probability as performing marginal MCMC on the space of model index.

The remaining difficulty is how to construct these transports in problems where the conditional posteriors themselves are hard to sample. The sample-trained TRJ implementation of \citet{C_AISTATS_davies2023transport} fits transport maps by minimizing a forward KL divergence using pilot samples from the conditional target distributions. Although effective when such draws are representative, this requires a preliminary posterior-simulation stage for each model. In difficult or large trans-dimensional spaces, the geometry that impedes RJMCMC can also make these pilot draws costly or unreliable, particularly for low-probability models. Moreover, the autoregressive flows used in that construction have a sequential inverse-sampling direction, which can become costly when a large number of proposal evaluations is required.

These limitations motivate a different map-fitting principle. We fit inverse transport maps from variational reference distributions to the conditional posteriors by reverse-KL variational inference with normalizing flows~\citep{C_ICML_rezende2015variational}. Thus, posterior draws are replaced by evaluations of the unnormalized posterior in the map-fitting objective. We use RealNVP rather than autoregressive flows because its coupling layers avoid coordinate-wise sequential inversion and allow the repeated transport evaluations required by RJMCMC proposals to be computed in parallel. A conditional saturated-space construction further trains model-specific maps jointly across dimensions.

The learned maps parameterize RJMCMC proposals, while the Metropolis--Hastings correction targets the exact trans-dimensional posterior in eq.~\eqref{eq:trans_model_posterior}. Reverse-KL training may under-cover separated modes or  tails~\citep{J_JASA_blei2017variational,C_ICML_rezende2015variational}; such approximation errors affect finite-time proposal efficiency, including acceptance and switching rates, but not the stationary target under the usual support and reversibility conditions. We characterize this effect through residual importance weights and assess both posterior model-probability accuracy and trans-dimensional mixing.

Our contributions are as follows.
\begin{itemize}
  \item We introduce an RJMCMC framework with transport proposals trained by
  reverse-KL variational inference. Map fitting uses reference draws and
  unnormalized conditional-target evaluations rather than pilot posterior
  samples, while the Metropolis--Hastings correction retains the exact
  trans-dimensional target.

  \item We develop a saturated-space conditional RealNVP construction that
  jointly learns model-specific transport maps across dimensions, shares
  information between models, and avoids the coordinate-wise sequential inverse
  evaluation of autoregressive flows.

  \item We derive a residual-weight representation of the trans-dimensional
  acceptance probability under approximate transports. It separates the
  finite-time loss of proposal efficiency caused by reverse-KL under-coverage
  from posterior validity under the usual support and reversibility conditions;
  a tempered reverse-KL objective stabilizes difficult conditional training.

  \item We evaluate the method on a controlled sinh--arcsinh example, Bayesian
  factor analysis, robust-regression variable selection, a coal-mining
  change-point model, and phylogenetic substitution-model selection on two
  sequence alignments. We report posterior accuracy, trans-dimensional mixing,
  and a separate accounting of RJMCMC sampling, pilot-sampling, and map-training
  costs.
\end{itemize}

The framework also yields within-model transported updates and evidence-related importance estimates as natural by-products. We discuss these extensions, but our main empirical comparisons focus on between-model proposal efficiency.

Code for numerical experiments is available at \href{https://github.com/Palantir-zoe/TRJ_VINFs}{TRJ-VINFs}.

The remainder of the paper is organized as follows. Section~\ref{sec:background} reviews RJMCMC, TRJ proposals, variational inference with normalizing flows, and the computational distinction between RealNVP and autoregressive flows. Section~\ref{sec:new-framework} introduces reverse-KL-trained transport proposals and the conditional trans-dimensional training construction. Section~\ref{sec:exmples} presents numerical experiments. Section~\ref{sec:discussion} concludes.

\section{Background}
\label{sec:background}

\subsection{Reversible Jump Markov Chain Monte Carlo}
\label{sec:rjmcmc}
To propose a move from $\boldsymbol{\omega} = (k, \boldsymbol{\theta}_k)$ to $\boldsymbol{\omega}' = (k',\boldsymbol{\theta}_{k'})$, where the dimensions of $\boldsymbol{\theta}_k$ and $\boldsymbol{\theta}_{k'}$ are $d_k$ and $d_{k'}$, RJMCMC first proposes a new model $k'$ with probability $q(k'|k)$. To solve the problem that $d_k$ and $d_{k'}$ may be different, RJMCMC utilizes \emph{dimension matching}, that is, introducing auxiliary variables $\boldsymbol{u}_k\sim\varphi_k$ and $\boldsymbol{u}_{k'}\sim \varphi_{k'}$ of dimensions $m_k$ and $m_{k'}$, such that $d_k + m_k=d_{k'} + m_{k'}$. Then a proposal of $\boldsymbol{\theta}_{k'}$ is generated by using $\boldsymbol{u}_k$, $\boldsymbol{u}_{k'}$ and a diffeomorphism $g_{k,k'}$ so that $(\boldsymbol{\theta}_{k'}, \boldsymbol{u}_{k'}) = g_{k,k'}(\boldsymbol{\theta}_k, \boldsymbol{u}_k)$. The new proposal $(k',\boldsymbol{\theta}_{k'})$ is accepted with probability
\begin{equation}
\alpha(\boldsymbol{\omega},\boldsymbol{\omega}') =
\min \left\{ 1,
\frac{\pi(\boldsymbol{\omega}'| \mathbf{y}) \varphi_{k'}(\boldsymbol{u}_{k'})
  q(k | k')}
{\pi(\boldsymbol{\omega}|\mathbf{y})\varphi_k(\boldsymbol{u}_k) q(k' | k)}
|J_{k,k'}| \right\},
\label{eq:rjmcmc_accept_prob}
\end{equation}
where $J_{k,k'}=\frac{\partial g_{k , k'}\left(\boldsymbol{\theta}_k, \boldsymbol{u}_k\right)}{\partial\left(\boldsymbol{\theta}_k, \boldsymbol{u}_k\right)}$ is the Jacobian matrix and $|\cdot|$ the absolute value of determinant. The choices of $\varphi_k$, $\varphi_{k'}$ and $g_{k,k'}$ have large impact on the efficiency of RJMCMC.

\subsection{Transport Reversible Jump Proposals}
\label{sec:trj}

For model $k$, if $T_k: \mathbb{R}^{d_k}\to \mathcal{Z}^{d_k}$ is a TM from $\pi_k$ to $\otimes_{d_k}\nu$, i.e., a reference distribution defined on $\mathcal{Z}^{d_k}$, then $T_k \# \pi_k = \otimes_{d_k}\nu$. Note that $\otimes_{d_k}\nu$ is the joint density of $d_k$ independent $\nu$-distributed variables, with $\nu$ to be some univariate distribution defined on $\mathcal{Z}\subset \mathbb{R}$.

TRJ utilizes the TMs $\{T_k\}$ to make trans-model proposals. Given (approximated) TMs for all the models, TRJ proposes a move from $k$ to $k'$ by  first transporting the current parameters $\boldsymbol{\theta}_k$ in model $k$ to the reference space, i.e.,
\begin{equation}
\boldsymbol{z}_k = T_k(\boldsymbol{\theta}_k).
\label{eq:theta-to-z}
\end{equation}
If $d_{k'} \ge d_k$, an auxiliary variable $\boldsymbol{u}_k$ of dimension $d_{k'}-d_k$ is drawn from $\otimes_{d_{k'}-d_k}\nu $, and a diffeomorphism $\bar{h}_{k,k'}$ is applied on the concatenated vector of $\boldsymbol{z}_k$ and $\boldsymbol{u}_k$ to obtain a reference variable $\boldsymbol{z}_{k'}$ corresponding to model $k'$, that is,
\begin{equation}
\boldsymbol{z}_{k'} = \bar{h}_{k,k'}(\boldsymbol{z}_k, \boldsymbol{u}_k).
\label{eq:h-reference}
\end{equation}
If $d_{k'} < d_k$, $\boldsymbol{u}_{k'}$ of dimension $d_k-d_{k'}$ is discarded from $\boldsymbol{z}_k$ before applying $\bar{h}_{k,k'}$. Note that $\bar{h}_{k,k'}$ should satisfy the pushforward-invariance and volume-preserving conditions and a default choice is the identity map \citep{C_AISTATS_davies2023transport}. Finally, a parameter value in model $k'$ is proposed by
\begin{equation}
\boldsymbol{\theta}_{k'} = T_{k'}^{-1}(\boldsymbol{z}_{k'}).
\label{eq:z-to-theta}
\end{equation}
\citet{C_AISTATS_davies2023transport} proved that if all the TMs are exact, then the TRJ proposals are in some sense optimal, with the acceptance probability in eq.~\eqref{eq:rjmcmc_accept_prob} reduced to:
\begin{equation}
\alpha(\boldsymbol{\omega}, \boldsymbol{\omega}') = \min \left\{1, \frac{\pi(k'| \mathbf{y}) q(k | k^{\prime})}{\pi(k | \mathbf{y}) q( k^{\prime} | k)}\right\},
\label{eq:accept_prob_simple}
\end{equation}
which depends on model index alone.

\citet{C_AISTATS_davies2023transport} used several chained AF models (details in Section~\ref{sec:nflow}) to approximate TMs, and trained them by minimizing the forward Kullback-Leibler (KL) divergence between the AF model and an empirical distribution estimated using the pilot samples from the target. The approximate TMs can be fitted for each model individually, but the computational cost is high if the number of models $K$ is large. To solve this problem, \citet{C_AISTATS_davies2023transport} introduced the \emph{conditional transport proposal} (CTP), which can obtain all the approximate TMs by training just once. Specifically, CTP considers the \emph{saturated-space} approach \citep{J_JRSSSB_brooks2003efficient} and augments the state space of each model to the maximum dimension $d_{max} = \max_{k\in\{1, 2,\dots,K\}}\{d_k\}$. The augmented target distribution is
\begin{equation}
 \tilde{\pi}(\tilde{\boldsymbol{\omega}}) = \pi(\boldsymbol{\omega} | \mathbf{y})
 (\otimes_{d_{max}-d_k}\nu)(\boldsymbol{u}_{\sim k}),
\end{equation}
where $\boldsymbol{\omega} = (k, \boldsymbol{\theta}_k)$, $\tilde{\boldsymbol{\omega}} = (k, \boldsymbol{\theta}_k, \boldsymbol{u}_{\sim k})$ and the auxiliary variable $\boldsymbol{u}_{\sim k}$ of dimension $d_{max}-d_k$ is drawn from a reference distribution $\otimes_{d_{max}-d_k}\nu$. In this setting,
all the approximate TMs $\{\tilde{T}(\cdot | k)\}$ can be learned by training one single \emph{conditional} NF given the model index $k$ as an input (context). The proposal from model $k$ to $k'$ is constructed by first drawing $k'$ from $q(\cdot | k)$ and then letting
$(\boldsymbol{\theta}_{k'},\boldsymbol{u}_{\sim k'})
=c_{k'}^{-1}\circ \tilde{T}^{-1}(\cdot |k')\circ\tilde{T}(\cdot|k)\circ c_k(\boldsymbol{\theta}_k,\boldsymbol{u}_{\sim k})$,
where ``$\circ$" represents composition, and $c_k$ is concatenation.

\subsection{Variational Inference with Normalizing Flows}
\label{sec:VI-with-NFs}
Variational inference (VI) approximates a posterior distribution by selecting
a member of a tractable family $\mathcal{Q}$ that minimizes the reverse KL
divergence~\citep{J_JASA_blei2017variational}:
\begin{equation}
q^*(\boldsymbol{\theta}) =
\arg\min_{q(\boldsymbol{\theta})\in\mathcal{Q}}
\operatorname{KL}\!\left[
q(\boldsymbol{\theta})\middle\|\pi(\boldsymbol{\theta}\mid\boldsymbol{y})
\right].
\label{eq:VI-KL}
\end{equation}
Because the marginal likelihood is generally unavailable, this is implemented
by maximizing the evidence lower bound
\begin{equation}
\operatorname{ELBO}(q)
=
\mathbb{E}_q[\log \pi(\boldsymbol{\theta},\boldsymbol{y})]
-
\mathbb{E}_q[\log q(\boldsymbol{\theta})],
\label{eq:elbo}
\end{equation}
since
$\operatorname{KL}(q\|\pi)=-\operatorname{ELBO}(q)+\log\pi(\boldsymbol{y})$.
Reverse-KL training requires samples from $q$ and evaluations of the
unnormalized target, but can under-cover separated modes or tails. Common
remedies include alternative divergences, tempering, and richer variational
families~\citep{C_NeurIPS_li2016renyi,C_NeurIPS_wang2018variational,
C_ICML_rezende2015variational,arXiv_qin2025flowvat,C_ICML_xu2023mixflows}.

We use finite normalizing flows to construct flexible members of
$\mathcal{Q}$~\citep{C_ICML_rezende2015variational}. Let
$\boldsymbol{\theta}^0\sim q_0$ be drawn from a tractable reference density
and define
$\boldsymbol{\theta}^{l}=f_l(\boldsymbol{\theta}^{l-1})$ for
$l=1,\ldots,L$, where each $f_l$ is invertible and differentiable. Then
\begin{equation}
\log q_L(\boldsymbol{\theta}^L)
=
\log q_0(\boldsymbol{\theta}^0)
-
\sum_{l=1}^{L}
\log\left|
\det J_{f_l}(\boldsymbol{\theta}^{l-1})
\right|,
\qquad
J_{f_l}(\boldsymbol{\theta}^{l-1})
=
\frac{\partial f_l(\boldsymbol{\theta}^{l-1})}
{\partial\boldsymbol{\theta}^{l-1}}.
\label{eq:nf-density}
\end{equation}
\subsection{RealNVP versus Autoregressive Flows}
\label{sec:nflow}

Recall that \citet{C_AISTATS_davies2023transport} specified each approximate transport map as a chain of AFs. One AF is defined element-wise as
\begin{equation}
T(\boldsymbol{z})_i
=
\tau\!\left(
z_i;\boldsymbol{\zeta}_i(\boldsymbol{z}_{<i};\boldsymbol{\psi})
\right),
\qquad i=1,\ldots,d,
\label{eq:AF-transform}
\end{equation}
where $\tau(\cdot;\boldsymbol{\xi})$ is a univariate diffeomorphism parametrized by $\boldsymbol{\xi}\in \boldsymbol{\Xi}$ with $\boldsymbol{\Xi}$ being the set of all admissible $\boldsymbol{\xi}$ values, and each of the functions $\{\boldsymbol{\zeta}_i: \mathbb{R}^{i-1}\to \boldsymbol{\Xi}\}$ is in practice the output of a single neural network that takes $\boldsymbol{z}_{<i}$ as input. This sequential inverse evaluation can be costly when transport maps are repeatedly used in RJMCMC proposals.

We instead use RealNVP~\citep{C_ICLR_dinh2017density}, which is built from \emph{affine coupling layers} (ACL). For $1\le d_0<d$, one such layer maps $\boldsymbol{z}$ to $\boldsymbol{x}$ by
\begin{equation}
\begin{aligned}
\boldsymbol{x}_{1:d_0}
&= \boldsymbol{z}_{1:d_0},\\
\boldsymbol{x}_{d_0+1:d}
&= \boldsymbol{z}_{d_0+1:d}\odot
\exp\!\left\{s(\boldsymbol{z}_{1:d_0})\right\}
+ t(\boldsymbol{z}_{1:d_0}),
\end{aligned}
\label{eq:realnvp-acl}
\end{equation}
where $s,t:\mathbb{R}^{d_0}\to\mathbb{R}^{d-d_0}$. Its Jacobian is block lower-triangular, with
$
\log\left|
\det\left(\frac{\partial\boldsymbol{x}}{\partial\boldsymbol{z}}\right)
\right|
=
\sum_{j=1}^{d-d_0}
s_j(\boldsymbol{z}_{1:d_0}).
$
Both directions of a coupling layer avoid coordinate-wise autoregressive recursion, so all coordinates in the transformed block can be evaluated in parallel. Alternating masks or permutations across layers ensure that all coordinates are transformed. This parallelism motivates our use of RealNVP for repeated trans-dimensional proposals.
\section{New Framework Using VI with NFs to Train TM-based RJMCMC Proposals}
\label{sec:new-framework}

We use VI with normalizing flows to train the transport maps used in TRJ proposals.  The fitted maps are then embedded in the usual Metropolis--Hastings-corrected RJMCMC transition, so they are proposal-calibration rather than replacements for posterior simulation.

\subsection{TRJ Proposals Trained by VI with Individual NFs}
\label{sec:trj-individual}
For model $k$, \citet{C_AISTATS_davies2023transport} specified $T_k$, the approximate TM from the target $\pi_k$ to the reference $\otimes_{d_k}\nu$, as AFs and fit it by minimizing the forward KL divergence, i.e.,
\begin{equation}\label{eq:forward-kl}
\text{KL}\left[\pi(\boldsymbol{\theta}_k|k, \boldsymbol{y})|| q(\boldsymbol{\theta}_k)\right] = \mathbb{E}_{\pi_k}\left[\log
\frac{\pi(\boldsymbol{\theta}_k|k, \boldsymbol{y})}{q(\boldsymbol{\theta}_k)}\right],
\end{equation}
where $q(\boldsymbol{\theta}_k) = (\otimes_{d_k}\nu)\left(T_k(\boldsymbol{\theta}_k)\right)\left|J_{T_k}(\boldsymbol{\theta}_k)\right|$. The KL divergence in eq.~\eqref{eq:forward-kl} is an expectation over the conditional target $\pi_k$ so that samples from $\pi_k$ are required for training.

We instead learn $T_k^{-1}$, i.e., TM from the reference to the target, using VI, which minimizes the reverse KL divergence, so that only requiring samples from the reference. We set $T_k^{-1}$ to be RealNVP with parameters $\boldsymbol{\eta}$, denoted by $f_{\boldsymbol\eta}$, which is a chain of ACLs as in eq.~\eqref{eq:realnvp-acl}. If $\boldsymbol{z}_k\sim\otimes_{d_k}\nu$, then $\boldsymbol{z}_{\boldsymbol\eta}= f_{\boldsymbol\eta}(\boldsymbol{z}_k)\sim q_{\boldsymbol\eta}$, where $q_{\boldsymbol\eta}\left(\boldsymbol{z}_{\boldsymbol\eta}\right) =  (\otimes_{d_k}\nu)(\boldsymbol{z}_k)\left|J_{f_{\boldsymbol\eta}} (\boldsymbol{z}_k)\right|^{-1}$. The VI with NFs minimizes
\begin{equation}\label{eq:reverse-kl}
\text{KL} \left[q_{\boldsymbol\eta}\left(\boldsymbol{z}_{\boldsymbol\eta}\right)|| \pi(\boldsymbol{z}_{\boldsymbol\eta}|k, \boldsymbol{y}) \right] = \mathbb{E}_{q_{\boldsymbol\eta}}\left[\log
\frac{q_{\boldsymbol\eta}\left(\boldsymbol{z}_{\boldsymbol\eta}\right)}{\pi(\boldsymbol{z}_{\boldsymbol\eta}|k, \boldsymbol{y})}\right],
\end{equation}
which is the reverse KL divergence between $q_{\boldsymbol\eta}$ and $\pi_k$. Recall that VI minimizes eq.~\eqref{eq:reverse-kl} by minimizing the negative ELBO, i.e., $\mathbb{E}_{q_{\boldsymbol\eta}}\left[\log
\frac{q_{\boldsymbol\eta}\left(\boldsymbol{z}_{\boldsymbol\eta}\right)}{\pi(\boldsymbol{z}_{\boldsymbol\eta}, \mathbf{y} | k)}\right]$, where $\pi(\boldsymbol{z}_{\boldsymbol\eta}, \mathbf{y} | k)=\pi(\boldsymbol{z}_{\boldsymbol\eta} | k) \pi(\mathbf{y} | \boldsymbol{z}_{\boldsymbol\eta}, k)$ is the un-normalized version of $\pi(\boldsymbol{z}_{\boldsymbol\eta}|\mathbf{y}, k)$. By applying the reparameterisation trick \citep{C_ICLR_kingma2014autoencoding}, the negative ELBO becomes
\begin{equation}\label{eq:elbo-repara}
\ell(\boldsymbol{\eta}) = \mathbb{E}_{\otimes_{d_k}\nu}\left[\log
\frac{q_{\boldsymbol\eta}\left(f_{\boldsymbol\eta}(\boldsymbol{z}_k)\right)}{\pi\left(f_{\boldsymbol\eta}(\boldsymbol{z}_k), \mathbf{y} | k\right)}\right],
\end{equation}
which is an expectation over the simple reference distribution $\otimes_{d_k}\nu$. We use stochastic gradient variational inference (SGVI) to optimize eq.~\eqref{eq:elbo-repara}; the corresponding fitting routine is given in Supplementary Appendix~B. The negative ELBO and its gradient are estimated with samples from $\otimes_{d_k}\nu$. For the numerical studies in this paper, we use ADAM~\citep{C_ICLR_kingma2015adam} as the optimizer.

In this paper, we set $\nu$ to be standard Gaussian $\mathcal{N}(0,1)$ so that $\otimes_{d_k}\nu$ is $\mathcal{N}_d(\boldsymbol{0},\boldsymbol{I})$. We also include $t$ distribution~\cite{J_MLST_andrade2024stabilizing} into our code as an option for $\nu$, which may be more suitable than Gaussian if the target is heavy-tailed.

\citet{C_AISTATS_davies2023transport} used the AF model to specify the TM $T_k$, and set the function $\tau(\cdot;\boldsymbol{\xi})$ in eq.~\eqref{eq:AF-transform} to be masked autoregressive transforms with rational quadratic splines \citep[RQMA,][]{D_durkan2020nflows} so that the model is tractable in both forward and reverse directions. However, the inverse sampling from $\boldsymbol{z_k}$ to $\boldsymbol{\theta_k}$ using the AF is sequential, with computational cost $\mathcal{O}(d_k)$, and not parallelizable. We instead specify $T_k^{-1}$ with RealNVP, which is tractable in both directions, but much more efficient in computation. For either direction, the evaluation of RealNVP is highly parallelisable, scaling well with the dimension. \citet{J_MLST_andrade2024stabilizing} showed in their numerical studies that evaluating the neural spline flows with 16 layers is more expensive than computing RealNVP with 64 ACLs. RealNVP is also powerful in representation. Under the conditions of \citet{C_ICML_koehler2021representational}, three alternating ACLs can approximate a bounded-support absolutely continuous target arbitrarily well in Wasserstein distance, demonstrating the
representational power of shallow RealNVP architectures. We specify the functions $s$ and $t$ in ACL to be multi-layer perceptrons (MLP) and set the depth of each RealNVP to be $L\ge 8$ in numerical examples. Although focusing on RealNVP, we integrate other commonly used flow models into our framework for users to choose, including masked autoregressive flows~\cite{C_NeurIPS_papamakarios2017masked}, neural spline flows \cite{D_durkan2020nflows}, planner flows \cite{C_ICML_rezende2015variational}, etc.

After obtaining $T_k^{-1}$ and $T_k$ for each model $k$ using the VI method described above, our algorithm proceeds identically to TRJ. Given the current state $\boldsymbol{\omega} = (k,\boldsymbol{\theta}_k)$, we sample $k'$ from $q(\cdot | k)$, and generate $\boldsymbol{\theta}_{k'}$ via the steps in eqs.~\eqref{eq:theta-to-z}--\eqref{eq:z-to-theta}. We set $\bar{h}_{k,k'}$ to be the identity map. The new proposal $\boldsymbol{\omega}' = (k',\boldsymbol{\theta}_{k'})$ is accepted with probability
\begin{equation}
\alpha(\boldsymbol{\omega},\boldsymbol{\omega}') =
\min \left\{ 1,
\frac{\pi(\boldsymbol{\omega}'\mid \mathbf{y})g'_u
  q(k | k')|J_{T_k}|}
{\pi(\boldsymbol{\omega}\mid \mathbf{y})g_u q(k' | k)|J_{T_{k'}}|} \right\},
\label{eq:accept_prob_vi}
\end{equation}
where $g_u = (\otimes_{d_{k'}-d_k}\nu)(\boldsymbol{u}_k)$, $g'_u = 1$ if $d_{k'} \ge d_k$, and $g_u = 1$, $g'_u = (\otimes_{d_k-d_{k'}}\nu)(\boldsymbol{u}_{k'})$ otherwise.

\subsection{New Trans-dimensional VI Method with Conditional NFs for TRJ Proposals}
\label{sec:ctp-vi}

Recall that \citet{C_AISTATS_davies2023transport} introduced CTP to learn the approximate TMs for all models by training a single conditional AF model. To achieve this by VI, we extend RealNVP to a conditional version and develop a trans-dimensional VI method with conditional NFs.

As in \citet{C_AISTATS_davies2023transport}, we adopt the saturated-space approach. For each model $k$, we use an auxiliary variable $\boldsymbol{u}_{\sim k}$ from the reference distribution $\otimes_{d_{\max}-d_k}\nu$ to augment the state space, so that the conditional target becomes
\begin{equation}
\tilde{\pi}(\tilde{\boldsymbol{\theta}}_k\mid\mathbf{y},k)
\propto
\pi(\mathbf{y}\mid\boldsymbol{\theta}_k,k)
\pi(\boldsymbol{\theta}_k\mid k)
(\otimes_{d_{\max}-d_k}\nu)(\boldsymbol{u}_{\sim k}),
\label{eq:aug-target-conditional}
\end{equation}
where $\tilde{\boldsymbol{\theta}}_k=
(\boldsymbol{\theta}_k,\boldsymbol{u}_{\sim k})$ has dimension $d_{\max}$. Marginalizing $\boldsymbol{u}_{\sim k}$ recovers the original conditional posterior $\pi(\boldsymbol{\theta}_k\mid\mathbf{y},k)$.

For flow training, let $\rho(k)>0$ be a user-specified distribution over model contexts. It determines how training effort is allocated across models, but is not a posterior model probability and is not part of the target of the final RJMCMC sampler. We use the uniform distribution on $\{1,\ldots,K\}$ for $\rho$ in the reported conditional-flow fits.

We specify a conditional variational distribution $\tilde{q}_{\boldsymbol{\eta},\boldsymbol{\phi}} (\tilde{\boldsymbol{\theta}}_k\mid k)$, where $\boldsymbol{\eta}$ denotes the conditional-flow parameters and $\boldsymbol{\phi}$ denotes the parameters of the conditional base distribution. For each affine coupling layer,
\begin{equation}
\begin{array}{l}
\boldsymbol{x}_{1:d_0}=\boldsymbol{z}_{1:d_0},\\
\boldsymbol{x}_{d_0+1:d_{\max}}
=
\boldsymbol{z}_{d_0+1:d_{\max}\!}\odot
\exp\bigl(s(\boldsymbol{z}_{1:d_0},\boldsymbol{c}_k)\bigr)
+
t(\boldsymbol{z}_{1:d_0},\boldsymbol{c}_k),
\end{array}
\label{eq:conditional-acl}
\end{equation}
where $\boldsymbol{c}_k$ is the context encoding of model $k$. Thus $s$ and $t$ are MLPs taking both $\boldsymbol{z}_{1:d_0}$ and $\boldsymbol{c}_k$ as inputs.

Let $\tilde{\boldsymbol{z}}_0$ follow the conditional base distribution
\begin{equation}
\tilde{q}_{0,\boldsymbol{\phi}}(\cdot\mid k)
=
\mathcal{N}_{d_{\max}}\left(
\boldsymbol{\mu}_{\boldsymbol{\phi}}(k),
\boldsymbol{\Sigma}_{\boldsymbol{\phi}}(k)
\right),
\end{equation}
where the mean vector and diagonal covariance matrix are outputs of MLPs with context $\boldsymbol{c}_k$. Applying the conditional RealNVP $\tilde{f}_{\boldsymbol{\eta}}(\cdot\mid\boldsymbol{c}_k)$ gives
\begin{equation}
\tilde{\boldsymbol{z}}_{\boldsymbol{\eta}}
=
\tilde{f}_{\boldsymbol{\eta}}
(\tilde{\boldsymbol{z}}_0\mid\boldsymbol{c}_k)
=
\tilde{f}_{L}\circ\cdots\circ\tilde{f}_{1}
(\tilde{\boldsymbol{z}}_0\mid\boldsymbol{c}_k).
\end{equation}
Its induced density is
\begin{equation}
\tilde{q}_{\boldsymbol{\eta},\boldsymbol{\phi}}
(\tilde{\boldsymbol{z}}_{\boldsymbol{\eta}}\mid k)
=
\tilde{q}_{0,\boldsymbol{\phi}}
\left(
\tilde{f}_{\boldsymbol{\eta}}^{-1}
(\tilde{\boldsymbol{z}}_{\boldsymbol{\eta}}
\mid\boldsymbol{c}_k)
\mid k
\right)
\left|
\det J_{\tilde{f}_{\boldsymbol{\eta}}^{-1}}
(\tilde{\boldsymbol{z}}_{\boldsymbol{\eta}}
\mid\boldsymbol{c}_k)
\right|.
\end{equation}

The conditional reverse-KL objective is the $\rho$-weighted sum of conditional negative ELBOs,
\begin{equation}
\begin{aligned}
\mathcal{L}_{\mathrm{CTP}}
(\boldsymbol{\eta},\boldsymbol{\phi})
&=
\mathbb{E}_{k\sim\rho}
\mathbb{E}_{\tilde{q}_{0,\boldsymbol{\phi}}
(\tilde{\boldsymbol{z}}_0\mid k)}
\left[
\log
\frac{
\tilde{q}_{\boldsymbol{\eta},\boldsymbol{\phi}}
\left(
\tilde{f}_{\boldsymbol{\eta}}
(\tilde{\boldsymbol{z}}_0\mid\boldsymbol{c}_k)
\mid k
\right)
}{
\tilde{\pi}\left(
\tilde{f}_{\boldsymbol{\eta}}
(\tilde{\boldsymbol{z}}_0\mid\boldsymbol{c}_k),
\mathbf{y}\mid k
\right)
}
\right].
\end{aligned}
\label{eq:elbo-conditional}
\end{equation}
Unknown conditional normalizing constants are additive terms independent of
$(\boldsymbol{\eta},\boldsymbol{\phi})$ and therefore do not affect this
optimization. The objective is estimated by reparameterization using samples
from $\tilde{q}_{0,\boldsymbol{\phi}}(\cdot\mid k)$; it does not require
samples from $\pi(\boldsymbol{\theta}_k\mid\mathbf{y},k)$.

For difficult conditional targets, the reverse-KL objective in
eq.~\eqref{eq:elbo-conditional} can be practically mode-seeking during
optimization. CTP therefore additionally permits a beta-conditioned tempered
training objective. Let $\tilde{\gamma}_k(\tilde{\boldsymbol{\theta}}_k)$
denote the unnormalized density on the right-hand side of
eq.~\eqref{eq:aug-target-conditional}, and let
$r_k(\tilde{\boldsymbol{\theta}}_k)$ be a tractable reference density on the
same saturated space. For $\beta\in[\beta_{\min},1]$, where
$0<\beta_{\min}\le 1$, define
\begin{equation}
\log \tilde{\gamma}_{k,\beta}(\tilde{\boldsymbol{\theta}}_k)
=
\beta\log \tilde{\gamma}_k(\tilde{\boldsymbol{\theta}}_k)
+
(1-\beta)\log r_k(\tilde{\boldsymbol{\theta}}_k).
\label{eq:tempered-ctp-target}
\end{equation}
The context is augmented to
$\boldsymbol{c}_{k,\beta}=(\boldsymbol{c}_k,\beta)$. Writing
$\tilde{q}_{\boldsymbol{\eta},\boldsymbol{\phi}}
(\cdot\mid\boldsymbol{c}_{k,\beta})$ for the same conditional flow and base
distribution evaluated at this augmented context, we minimize
\begin{equation}
\begin{aligned}
\mathcal{L}_{\mathrm{temp}}
(\boldsymbol{\eta},\boldsymbol{\phi})
&=
\mathbb{E}_{\beta\sim\mathcal{U}[\beta_{\min},1]}
\mathbb{E}_{k\sim\rho}
\mathbb{E}_{
\tilde{q}_{\boldsymbol{\eta},\boldsymbol{\phi}}
(\cdot\mid\boldsymbol{c}_{k,\beta})}
\left[
\log
\tilde{q}_{\boldsymbol{\eta},\boldsymbol{\phi}}
(\tilde{\boldsymbol{\theta}}_k\mid\boldsymbol{c}_{k,\beta})
-
\log
\tilde{\gamma}_{k,\beta}(\tilde{\boldsymbol{\theta}}_k)
\right].
\end{aligned}
\label{eq:tempered-ctp-loss}
\end{equation}
At $\beta=1$, the tempered target is the original augmented target.
Values $\beta<1$ provide intermediate targets that can stabilize training.
The final proposal evaluates
$\tilde{f}_{\boldsymbol{\eta}}(\cdot\mid\boldsymbol{c}_{k,1})$; the
Metropolis--Hastings correction below continues to use the untempered
augmented target. Thus, tempering changes the fitted proposal but not the
invariant distribution.

The corresponding conditional stochastic-gradient fitting procedure is given
in Supplementary Appendix~B.

For each sampled $k$, a binary mask
$\boldsymbol{\gamma}_k$ identifies the $d_k$ active coordinates of
$\tilde{\boldsymbol{z}}_{\boldsymbol{\eta}}$. These coordinates are used to
evaluate the unnormalized conditional posterior
$\pi(\cdot,\mathbf{y}\mid k)$, while the inactive coordinates are evaluated
under $\otimes_{d_{\max}-d_k}\nu$.

After fitting, the conditional flow supplies a model-specific map at the
$\beta=1$ context,
\[
\tilde{T}^{-1}(\cdot\mid\boldsymbol{c}_{k,1})
=
\tilde{f}_{\boldsymbol{\eta}}
(\cdot\mid\boldsymbol{c}_{k,1}).
\]
The final sampler nevertheless targets the true augmented posterior
\begin{equation}
\Pi^{\mathrm{aug}}
(\tilde{\boldsymbol{\theta}}_k,k\mid\mathbf{y})
\propto
\pi(k)
\pi(\mathbf{y}\mid\boldsymbol{\theta}_k,k)
\pi(\boldsymbol{\theta}_k\mid k)
(\otimes_{d_{\max}-d_k}\nu)(\boldsymbol{u}_{\sim k}),
\label{eq:augmented-target}
\end{equation}
rather than the training distribution $\rho$. Let $Q_{\boldsymbol{\eta}}$
denote the conditional proposal density induced by the fitted transformations,
the reference auxiliary variables, and their Jacobian terms. The CTP move is
accepted with probability
\begin{equation}
\alpha_{\mathrm{CTP}}
=
1\wedge
\frac{
\Pi^{\mathrm{aug}}
(\tilde{\boldsymbol{\theta}}_{k'}',k'\mid\mathbf{y})\,
q(k\mid k')\,
Q_{\boldsymbol{\eta}}
\{(k,\tilde{\boldsymbol{\theta}}_k)
\mid(k',\tilde{\boldsymbol{\theta}}_{k'}')\}
}{
\Pi^{\mathrm{aug}}
(\tilde{\boldsymbol{\theta}}_k,k\mid\mathbf{y})\,
q(k'\mid k)\,
Q_{\boldsymbol{\eta}}
\{(k',\tilde{\boldsymbol{\theta}}_{k'}')
\mid(k,\tilde{\boldsymbol{\theta}}_k)\}
}.
\label{eq:accept_prob_vi_conditional}
\end{equation}
Thus, conditional-flow approximation affects proposal efficiency but not the invariant distribution of the Metropolis--Hastings-corrected chain.
\paragraph{Additional by-products.}
Although the comparisons focus on between-model proposals, the fitted maps have two optional uses.  First, for a fixed model $k$, a conventional within-model kernel can be run in transported coordinates with target $\pi_k\{T_k^{-1}(\boldsymbol z)\}\lvert\det\nabla T_k^{-1}(\boldsymbol z)\rvert$, in the spirit of NeuTraHMC~\citep{arXiv_hoffman2019neutralizing}; the samplers instead retain their conventional problem-specific within-model kernels.  Second, the fitted density $q_{\boldsymbol\eta,k}$ can serve as an importance proposal for $Z_k=\pi(\mathbf y\mid k)$, through $\widehat Z_k=m^{-1}\sum_{i=1}^m\gamma_k(\boldsymbol\theta^i)/q_{\boldsymbol\eta,k}(\boldsymbol\theta^i)$ with $\boldsymbol\theta^i\sim q_{\boldsymbol\eta,k}$, subject to the usual support and moment conditions.  With exact maps and $q(k'\mid k)=\pi(k'\mid\mathbf y)$, TRJ attains the ideal rejection-free model-index move of \citet{C_AISTATS_davies2023transport}; reliable evidence estimates could in principle inform this proposal.  Because reverse-KL under-coverage can make importance estimates unstable, we do not use them to choose model-index proposals.

\subsection{Approximate reverse-KL transports and mode seeking}
\label{sec:approx_rkl_theory_revised}

The exact-map theory of TRJ describes the ideal case in which each conditional posterior is transported to its reference distribution $\otimes_{d_k}\nu$. Here the maps are fitted by reverse-KL variational inference and are approximate. For individual maps, the following factorization shows that map error affects finite-time efficiency through residual importance weights, while the Metropolis--Hastings correction preserves the RJMCMC invariant distribution. Conditional flows remain exactly corrected by \eqref{eq:accept_prob_vi_conditional}, but are not covered by this simplification.

Let $\mathcal K=\{1,\ldots,K\}$. For model $k$, write
$
  \pi_k(\theta) := \pi(\theta\mid k,\mathbf y).
$
Let $q(k'\mid k)$ be the probability of proposing model $k'$ from model $k$. Let $T_k:\Theta_k\to Z^{d_k}$ be the learned map from the parameter space to the reference space, and define the density induced on $\Theta_k$ by its inverse as
\begin{equation}
  q_k(\theta)
  :=
  (\otimes_{d_k}\nu)\{T_k(\theta)\}\,
  \left|\det \nabla T_k(\theta)\right|.
  \label{eq:qk_transport_density_revised}
\end{equation}
The residual importance weight of the learned map is
\begin{equation}
  w_k(\theta) := \frac{\pi_k(\theta)}{q_k(\theta)},
  \qquad
  \ell_k(\theta) := \log w_k(\theta).
  \label{eq:residual_weight_revised}
\end{equation}
If $T_k\#\pi_k=\otimes_{d_k}\nu$, then $q_k=\pi_k$ and $w_k\equiv 1$.

Assume that $\pi(k\mid\mathbf y)>0$ for the model pairs, each $T_k$ is a diffeomorphism on the relevant support, and the reference-space dimension-matching map is volume-preserving. Thus, for $d_{k'}\ge d_k$,
$
  (\bar h_{k,k'})_\#
  \left\{(\otimes_{d_k}\nu)
  \otimes(\otimes_{d_{k'}-d_k}\nu)\right\}
  =
  \otimes_{d_{k'}}\nu,
  \left|\det \nabla \bar h_{k,k'}\right|=1,
$
with the analogous condition for the inverse move.

\begin{proposition}[Residual-weight factorization]
\label{prop:residual_weight_factorization_revised}
For any learned diffeomorphisms $T_k$, not necessarily exact transports, the RJMCMC acceptance ratio for a proposal $(k,\theta)\to(k',\theta')$ constructed by the TRJ mechanism is
\begin{equation}
  R_T\{(k,\theta),(k',\theta')\}
  =
  R_\star(k,k')\,
  \frac{w_{k'}(\theta')}{w_k(\theta)},
  \label{eq:residual_factorization_revised}
\end{equation}
where
$
  R_\star(k,k')
  :=
  \frac{\pi(k'\mid\mathbf y)q(k\mid k')}
       {\pi(k\mid\mathbf y)q(k'\mid k)}.
$
Consequently,
\[
  \alpha_T\{(k,\theta),(k',\theta')\}
  =
  1\wedge
  \left[
    R_\star(k,k')
    \frac{w_{k'}(\theta')}{w_k(\theta)}
  \right].
\]
If $T_k\#\pi_k=\otimes_{d_k}\nu$ for all $k$, then $w_k\equiv 1$ and the exact-map TRJ acceptance probability $1\wedge R_\star(k,k')$ is recovered.
\end{proposition}

Thus, the learned maps define proposals rather than a substitute for posterior simulation: residual weights correct their approximation error in the Metropolis--Hastings ratio. A poor variational fit can lower acceptance or slow model switching, but does not change the stationary distribution under the stated support and reversibility conditions.

As a direct consequence, let
\begin{equation}
  \Delta_{k,k'}(\theta,\theta')
  :=
  \ell_{k'}(\theta')-\ell_k(\theta).
  \label{eq:residual_log_difference_revised}
\end{equation}
Then
\[
  \alpha_T\{(k,\theta),(k',\theta')\}
  =
  1\wedge\{R_\star(k,k')\exp(\Delta_{k,k'}(\theta,\theta'))\}.
\]
If $|\Delta_{k,k'}(\theta,\theta')|\le \eta$, then
\begin{equation}
  e^{-\eta}\{1\wedge R_\star(k,k')\}
  \le
  \alpha_T\{(k,\theta),(k',\theta')\}
  \le
  e^{\eta}\{1\wedge R_\star(k,k')\}.
  \label{eq:acceptance_bound_revised}
\end{equation}
In particular, $\eta=\epsilon_k+\epsilon_{k'}$ is valid when $|\ell_k(\theta)|\le\epsilon_k$ and $|\ell_{k'}(\theta')|\le\epsilon_{k'}$. Hence approximate TRJ is close to its exact-map counterpart whenever residual log-weights have small pairwise fluctuations on the regions visited by the chain.

\paragraph{Mode seeking as proposal under-coverage.}
Reverse-KL training can favor high-density regions and underrepresent separated modes or tails~\citep{J_JASA_blei2017variational,C_ICML_rezende2015variational}. In the present framework this is a proposal-quality issue: the flow is used only in an MH-corrected transition.

Adapting the effective-support diagnostic of
\citet{J_PRD_nicoli2023detecting} to the present conditional setting, define, for $\varepsilon\in(0,1)$,
\begin{equation}
  S_{k,\varepsilon}
  :=
  \{\theta\in\Theta_k:q_k(\theta)\le\varepsilon\pi_k(\theta)\},
  \qquad
  m_{k,\varepsilon}:=\pi_k(S_{k,\varepsilon}).
  \label{eq:dropped_set_revised}
\end{equation}
This is a relative under-coverage set. Directly from its definition,
\begin{equation}
  q_k(S_{k,\varepsilon})
  \le \varepsilon\,\pi_k(S_{k,\varepsilon})
  =\varepsilon m_{k,\varepsilon}.
  \label{eq:undercoverage_mass_bound_revised}
\end{equation}
Thus a posterior-relevant region may receive little proposal mass even when the learned map is accurate on the region it covers. This creates a finite-time entry bottleneck, but not asymptotic bias after Metropolis--Hastings correction. Further stationary-chain and sticky-exit bounds are given in the supplementary material.

Reverse-KL training also flattens the conditional posterior in the reference coordinates used by TRJ. For fixed $k$, define
$
  r_k(z)
  :=
  \pi_k(T_k^{-1}(z))
  \left|\det\nabla T_k^{-1}(z)\right|.
$
Then a change of variables gives
\begin{equation}
  \mathrm{KL}(q_k\|\pi_k)
  =
  \mathrm{KL}(\otimes_{d_k}\nu\|r_k).
  \label{eq:flattening_identity_revised}
\end{equation}
Thus the reverse-KL objective makes the transported posterior close to the reference law in the same divergence. It can improve the geometry of the covered posterior mass, whereas under-coverage can still limit finite-time exploration.

\section{Numerical Experiments}
\label{sec:exmples}
\paragraph{Evaluation criteria}
We assess posterior model-probability estimation and mixing. We display running estimates of selected posterior model probabilities and report final estimates together with $|\hat p-p_{\mathrm{ref}}|$ when a reference value is available. Cross-model mixing is summarized by jump acceptance, the proportion of proposed between-model moves that are accepted, and by the model-switching rate, the proportion of iterations in which the model index changes. We also report recorded RJMCMC sampling-phase times. For the sample-trained transport methods, $N$ denotes the number of training samples from each model-specific conditional target; related pilot-sampling and map-fitting times are given in Supplementary Appendix~D. We do not emphasize effective sample sizes for discrete model indicators, which can be unstable across short repeated chains with long model sojourns.

\subsection{Illustrative Example}
\label{sec:sas}
Following \citet{C_AISTATS_davies2023transport}, we first consider a two-model sinh--arcsinh (SAS) target for which the exact transport maps are available. 
Let the two model dimensions be $d_1=1$ and $d_2=2$. For a given dimension $d$, parameters $\boldsymbol{\epsilon}\in\mathbb{R}^d$ and $\boldsymbol{\delta}\in\mathbb{R}_+^d$, define the element-wise SAS transformation of \citet{J_BIOMET_jones2009sinharcsinh} by
$
S_{\boldsymbol{\epsilon},\boldsymbol{\delta}}(\boldsymbol{x})
=
\sinh\!\left\{
\boldsymbol{\delta}^{-1}\odot
\left(\sinh^{-1}(\boldsymbol{x})+\boldsymbol{\epsilon}\right)
\right\},
$
where $\odot$ denotes element-wise multiplication. Its inverse is
$
S_{\boldsymbol{\epsilon},\boldsymbol{\delta}}^{-1}(\boldsymbol{x})
=
\sinh\!\left\{
\boldsymbol{\delta}\odot\sinh^{-1}(\boldsymbol{x})-\boldsymbol{\epsilon}
\right\}.
$
If $\boldsymbol{z}\sim\mathcal{N}_d(\boldsymbol{0},\boldsymbol{I}_d)$ and $\boldsymbol{\theta}=S_{\boldsymbol{\epsilon},\boldsymbol{\delta}}(\boldsymbol{L}\boldsymbol{z})$, then the induced density is
$
\pi_{\boldsymbol{\epsilon},\boldsymbol{\delta},\boldsymbol{L}}(\boldsymbol{\theta})
=
\phi_{\boldsymbol{L}\boldsymbol{L}^{\top}}
\!\left(
S_{\boldsymbol{\epsilon},\boldsymbol{\delta}}^{-1}(\boldsymbol{\theta})
\right)
\,
\left|
J_{S_{\boldsymbol{\epsilon},\boldsymbol{\delta}}^{-1}}
(\boldsymbol{\theta})
\right|,
$
where
$\phi_{\boldsymbol{L}\boldsymbol{L}^{\top}}(\cdot)$ denotes the density of $\mathcal{N}_d(\boldsymbol{0},\boldsymbol{L}\boldsymbol{L}^{\top})$, and $J_{S_{\boldsymbol{\epsilon},\boldsymbol{\delta}}^{-1}}(\boldsymbol{\theta})$ is the Jacobian matrix of the inverse SAS transformation. We assign model probabilities $\Pr(k=1)=1/4$ and $\Pr(k=2)=3/4$, so that the joint trans-dimensional target is
\begin{equation}
\pi(k,\boldsymbol{\theta}_k)
=
\begin{cases}
\dfrac14\,
\pi_{\boldsymbol{\epsilon}_1,\boldsymbol{\delta}_1,\boldsymbol{L}_1}
(\boldsymbol{\theta}_1),
& k=1,\\[1.2ex]
\dfrac34\,
\pi_{\boldsymbol{\epsilon}_2,\boldsymbol{\delta}_2,\boldsymbol{L}_2}
(\boldsymbol{\theta}_2),
& k=2.
\end{cases}
\label{eq:sas-target}
\end{equation}
The model-specific parameters are
$\boldsymbol{\epsilon}_1=-2$, $\boldsymbol{\delta}_1=1$, $\boldsymbol{L}_1=1$,
and
$\boldsymbol{\epsilon}_2=(1.5,-2)$, $\boldsymbol{\delta}_2=(1,1.5)$, where $\boldsymbol{L}_2$ is lower triangular and satisfies
$
\boldsymbol{L}_2\boldsymbol{L}_2^\top
=
\begin{bmatrix}
1 & 0.99\\
0.99 & 1
\end{bmatrix}.
$

We compare four choices for the model-specific TRJ maps. The exact map is included as an oracle benchmark. The affine and RQMA maps are trained from $N=5\times 10^4$ exact samples per model, matching the setup of \citet{C_AISTATS_davies2023transport}. Our reverse-KL approach fits the maps without target samples, using standard normal bases and normalizing-flow maps fitted separately for each model. 

\begin{figure}[!tbp]
\centering
\begin{subfigure}[t]{0.46\linewidth}
    \centering
    \includegraphics[width=\linewidth]{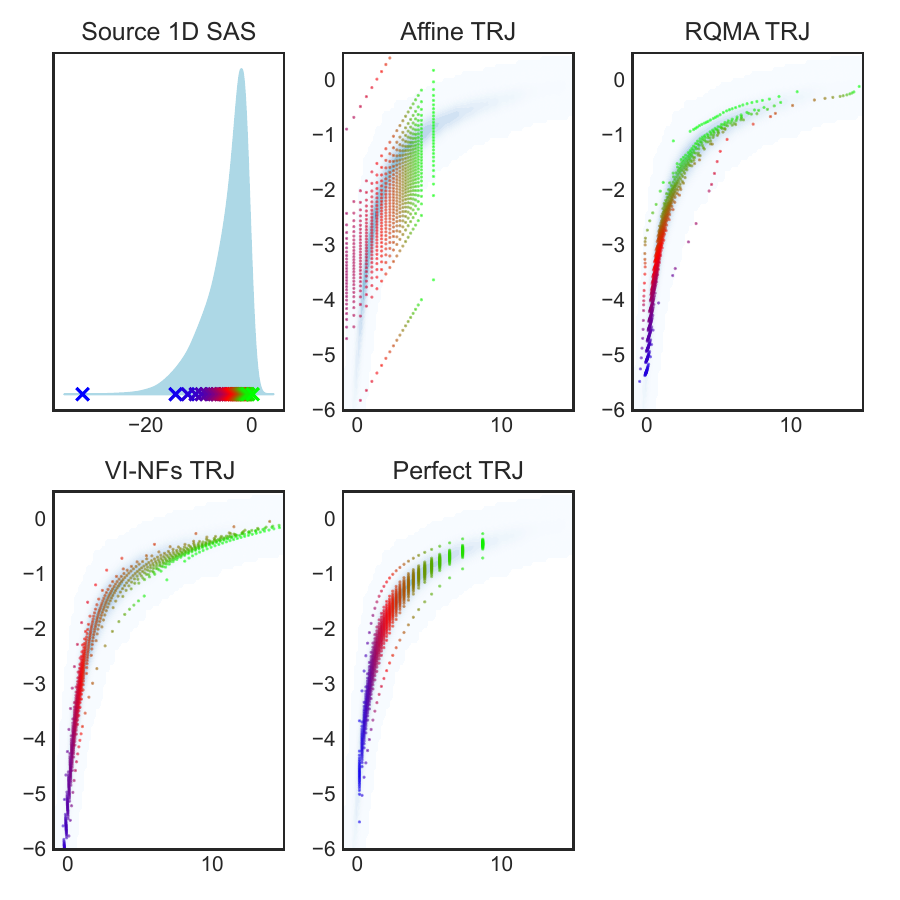}
    \caption{One-step proposal draws}
    \label{fig:sas-proposal}
\end{subfigure}\hfill
\begin{subfigure}[t]{0.50\linewidth}
    \centering
    \includegraphics[width=\linewidth]{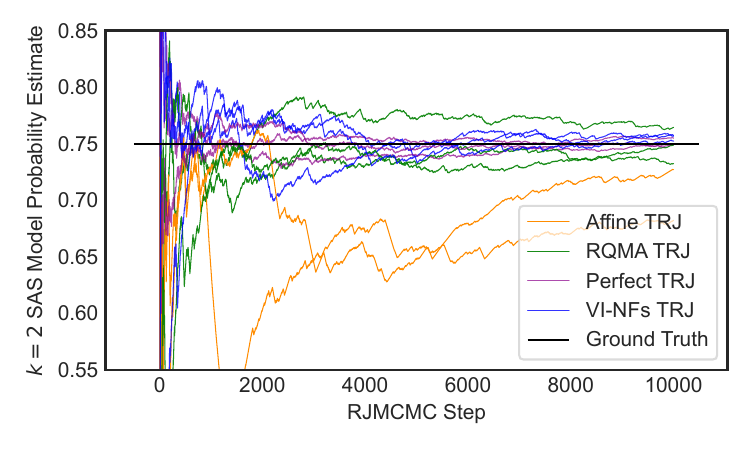}
    \caption{Running estimate of $\Pr(k=2)$}
    \label{fig:sas-running-mp}
\end{subfigure}

\caption{Illustrative SAS example. Panel (a) shows systematic draws from the
conditional target under model $k=1$ transported into the parameter space of
model $k=2$ using affine-map TRJ, RQMA-map TRJ, VI-NFs TRJ, and exact-map TRJ.
Panel (b) shows running estimates of $\Pr(k=2)$ under the target in
eq.~\eqref{eq:sas-target}.}
\label{fig:sas-proposal-draw}
\end{figure}

Figure~\ref{fig:sas-proposal-draw} summarizes both the one-step proposal behavior and the resulting RJMCMC estimates. Panel (a) compares proposals from model $k=1$ to model $k=2$: VI-NFs TRJ produces draws that are visually close to the oracle exact-map TRJ proposal, and closer than affine-map TRJ and RQMA-map TRJ trained from target samples. Panel (b) reports running estimates of $\Pr(k=2)$ over three independent RJMCMC chains targeting eq.~\eqref{eq:sas-target}, using the independent model proposal of \citet{C_AISTATS_davies2023transport} with probabilities $1/4$ and $3/4$ for models $1$ and $2$, respectively. The VI-NFs TRJ sampler closely tracks the oracle exact-map TRJ sampler and reaches the true probability $3/4$ substantially faster than the sample-trained affine-map TRJ and RQMA-map TRJ alternatives.

\subsection{Bayesian Factor Analysis}
\label{sec:fa}
We next revisit the Bayesian factor-analysis model-selection example of \citet{J_STATSIN_lopes2004bayesian}, also used by \citet{C_AISTATS_davies2023transport}. The data consist of monthly exchange rates of six currencies relative to the British pound, observed from January 1975 to December 1986. After standardization, this gives $n=143$ observations $\mathbf{y}_i\in\mathbb{R}^6$. Under a model with $k$ latent factors, $\mathbf{y}_i\mid \boldsymbol{\beta}_k,\boldsymbol{\Lambda}\sim\mathcal{N}_6(\mathbf{0},\boldsymbol{\beta}_k\boldsymbol{\beta}_k^\top+\boldsymbol{\Lambda})$ for $i=1,\ldots,143$, where $\boldsymbol{\Lambda}$ is a positive diagonal $6\times6$ matrix and $\boldsymbol{\beta}_k=[\beta_{ij}]$ is a $6\times k$ lower-triangular loading matrix with strictly positive diagonal entries. The model-specific parameter is $\boldsymbol{\theta}_k=(\boldsymbol{\beta}_k,\boldsymbol{\Lambda})$, with dimension $6(k+1)-k(k-1)/2$, and we compare the two- and three-factor models.

Following \citet{C_AISTATS_davies2023transport}, we assign independent priors
$$
\begin{aligned}
\beta_{ij} &\sim \mathcal{N}(0,1), && i>j,\\
\beta_{jj} &\sim \mathcal{N}_+(0,1), && j=1,\dots,k,\\
\Lambda_{ii} &\sim \mathcal{IG}(1.1,0.05), && i=1,\dots,6,
\end{aligned}
$$
where $\mathcal{N}_+(0,1)$ denotes the standard Gaussian distribution truncated to the positive real line and $\mathcal{IG}$ denotes the inverse-Gamma distribution. We assign equal prior probability to the two models, $\pi(k)=1/2$ for $k\in\{2,3\}$. The resulting joint posterior distribution is
\begin{equation}
\pi(k,\boldsymbol{\theta}_k\mid\mathbf{y})
\propto
\prod_{i=1}^{143}
\phi_{\boldsymbol{\beta}_k\boldsymbol{\beta}_k^\top+\boldsymbol{\Lambda}}
(\mathbf{y}_i)\,
\pi(\boldsymbol{\beta}_k\mid k)\,
\pi(\boldsymbol{\Lambda})\,
\pi(k),
\label{eq:fa-target}
\end{equation}
where $\mathbf{y}=(\mathbf{y}_1,\dots,\mathbf{y}_{143})$,
$\phi_{\boldsymbol{\beta}_k\boldsymbol{\beta}_k^\top+\boldsymbol{\Lambda}}(\cdot)$ denotes the density of the Gaussian distribution with covariance matrix $\boldsymbol{\beta}_k\boldsymbol{\beta}_k^\top+\boldsymbol{\Lambda}$, and
$\pi(\boldsymbol{\beta}_k\mid k)$ and $\pi(\boldsymbol{\Lambda})$ denote the priors defined above.

We compare four RJMCMC samplers for eq.~\eqref{eq:fa-target}, differing only in the between-model proposal. The first is the independent proposal of \citet{J_STATSIN_lopes2004bayesian}, with model-specific proposal parameters estimated from conditional posterior samples. The next two are the affine-map and RQMA-map TRJ proposals of \citet{C_AISTATS_davies2023transport}, again fitted from samples of the conditional targets. The fourth is VI-NFs TRJ, which uses standard normal bases and RealNVP maps fitted by reverse-KL VI as in Section~\ref{sec:trj-individual}. Positivity constraints on the diagonal loadings and uniqueness variances are handled using softplus transformations~\citep{J_JMLR_kucukelbir2017automatic}.

The first three proposals use pilot samples from the model-specific posterior distributions to fit their proposal maps. Following \citet{C_AISTATS_davies2023transport}, we generate these samples using static temperature-annealed SMC~\citep{J_JRSSSB_moral2006sequential} for the two-factor model and NUTS~\citep{J_JMLR_hoffman2014nouturn}, implemented in PyMC~\citep{J_PEERJCS_salvatier2016probabilistic}, for the three-factor model.  

\begin{figure}[!tbp]
\centering
\begin{subfigure}[t]{0.60\linewidth}
  \centering
  \includegraphics[width=\linewidth]{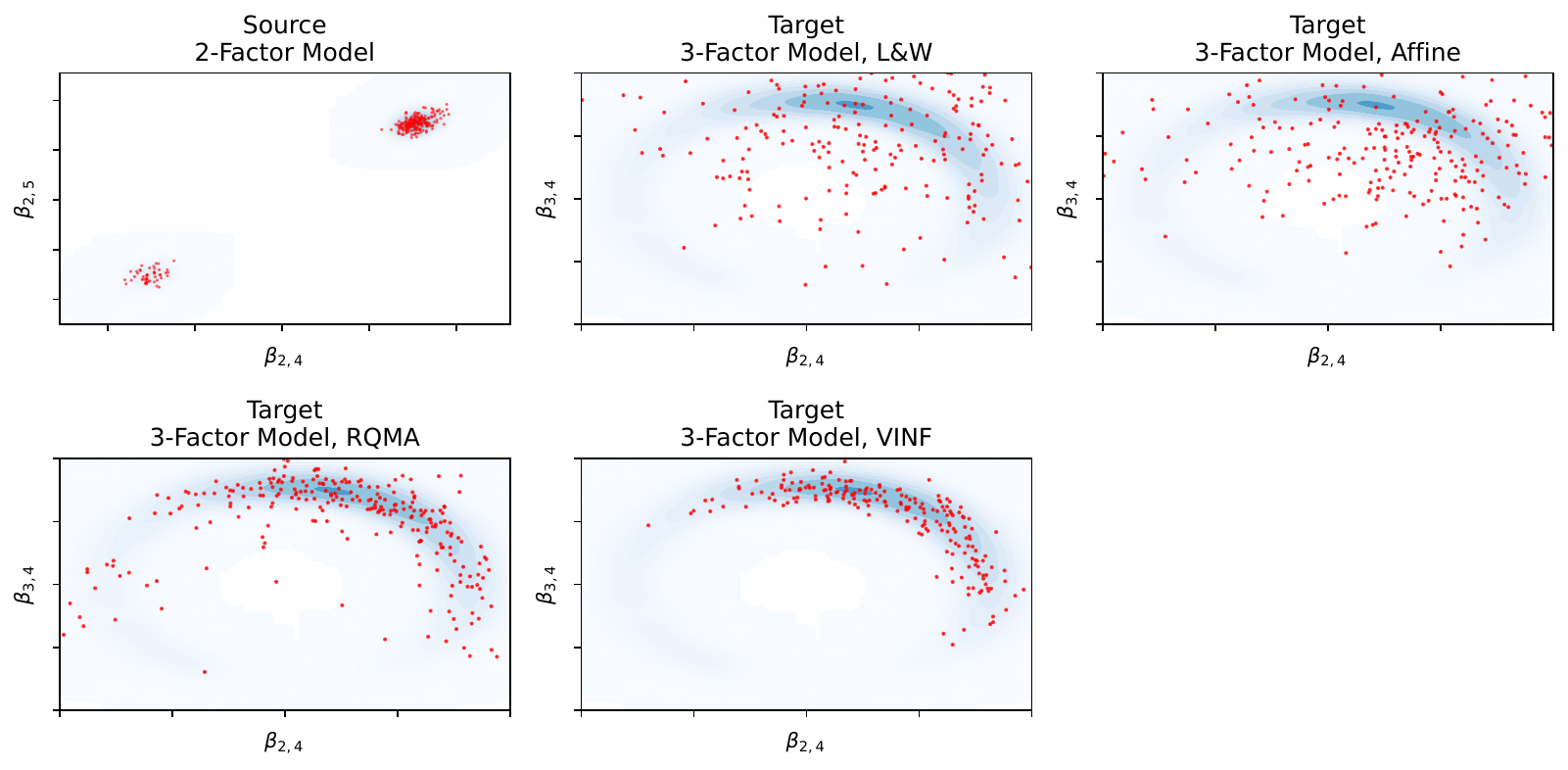}
  \caption{One-step two- to three-factor proposals}
  \label{fig:fa-proposal-compare}
\end{subfigure}\hfill
\begin{subfigure}[t]{0.36\linewidth}
  \centering
  \includegraphics[width=\linewidth]{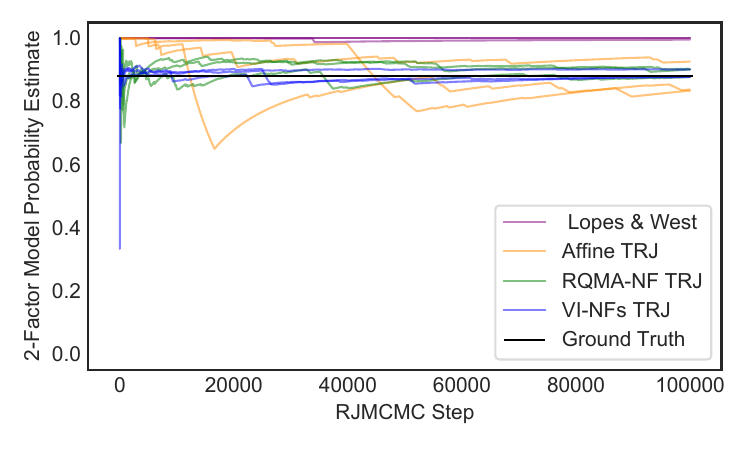}
  \caption{Running estimate of $\Pr(k=2\mid\mathbf y)$}
  \label{fig:fa-running-mp}
\end{subfigure}
\caption{Factor-analysis proposal and sampling diagnostics. Panel (a) shows systematic draws from the two-factor conditional target (red, left) and their one-step proposals for the three-factor model (red, remaining panels); blue contours show the corresponding conditional targets. Panel (b) shows running estimates of the two-factor posterior probability under the Lopes--West, affine-map TRJ, RQMA-map TRJ, and VI-NFs TRJ samplers. Three chains are run for each sampler, and the reference probability is $0.88$.}
\label{fig:fa-proposal-and-running}
\end{figure}

Figure~\ref{fig:fa-proposal-and-running}(a) gives a two-coordinate proposal diagnostic for a two- to three-factor move.  Relative to the Lopes--West and affine proposals, the VI-NFs map places the transported draws more consistently along a high-density region of the three-factor conditional target; the sample-trained RQMA map gives a similarly concentrated proposal.

\begin{table}[!tbp]
\centering
\caption{Sampling and model-probability diagnostics for the factor-analysis example at $N=4000$. The wall-clock column is the recorded 100,000-iteration RJMCMC sampling phase in seconds, reported as the mean $\pm$ sample standard deviation across three runs. Here $p_{\mathrm{ref}}=0.88$ is the two-factor posterior probability reported by \citet{J_STATSIN_lopes2004bayesian}. }
\label{tab:fa-total-time}
\resizebox{\linewidth}{!}{%
\begin{tabular}{lccccc}
\toprule
Proposal & RJMCMC sampling (s) & $\hat p(k=2)$ &
$|\hat p-p_{\mathrm{ref}}|$ & Jump accept. & Switch \\
\midrule
Lopes--West & $976.5\pm5.7$ & 0.5119 & 0.3681 &
$2.5\times10^{-6}$ & $1.2\times10^{-6}$ \\
Affine TRJ & $1021.1\pm6.5$ & 0.8484 & 0.0316 &
0.0031 & 0.0015 \\
RQMA-NF TRJ & $5633.1\pm36.3$ & 0.8990 & 0.0190 &
0.0666 & 0.0333 \\
VI-NFs TRJ & $971.3\pm1.2$ & 0.8648 & 0.0152 &
0.1800 & 0.0900 \\
\bottomrule
\end{tabular}%
}
\end{table}

Table~\ref{tab:fa-total-time} shows that, VI-NFs TRJ has the largest jump-acceptance and switching diagnostics, while its recorded sampling time is comparable with the Lopes--West and affine proposals and far below that of RQMA-NF TRJ. Appendix~D reports the pilot-draw and map-construction components separately. Figure~\ref{fig:fa-proposal-and-running}(b) compares running estimates of the two-factor posterior probability with the reference value 0.88 reported by \citet{J_STATSIN_lopes2004bayesian}.

\subsection{Variable Selection in Robust Regression Analysis}
\label{sec:vs}

We consider the robust regression variable-selection example of \citet{C_AISTATS_davies2023transport}. Let $\mathbf{x}_i\in\mathbb{R}^4$ denote the covariate vector for observation $i$, including an intercept, and let $\boldsymbol{\beta}=(\beta_0,\ldots,\beta_3)$ be the full regression coefficient vector. Conditional on the active coefficients, the response is modelled using the robust Gaussian mixture
$$
y_i\mid \boldsymbol{\beta},\boldsymbol{\gamma}
\sim
\frac12\mathcal{N}\!\left(\mathbf{x}_i^\top\{\boldsymbol{\beta}\odot\mathbf{m}(\boldsymbol{\gamma})\},1\right)
+
\frac12\mathcal{N}\!\left(\mathbf{x}_i^\top\{\boldsymbol{\beta}\odot\mathbf{m}(\boldsymbol{\gamma})\},100\right),
\qquad i=1,\ldots,80,
$$
where $\boldsymbol{\gamma}$ is a block-inclusion indicator and $\odot$ denotes element-wise multiplication. Following \citet{C_AISTATS_davies2023transport}, our primary four-model experiment uses the block indicator
$$
\boldsymbol{\gamma}=(1,\gamma_1,\gamma_2)\in\{1\}\times\{0,1\}^2,
\qquad
\mathbf{m}(\boldsymbol{\gamma})=(1,\gamma_1,\gamma_2,\gamma_2).
$$
Thus $\beta_0$ is always included, $\beta_1$ is controlled by $\gamma_1$, and $(\beta_2,\beta_3)$ are included or excluded jointly through $\gamma_2$. The model space has four elements, the block sizes are $(1,1,2)$, and the active parameter dimension is $d_{\boldsymbol{\gamma}}=1+\gamma_1+2\gamma_2$. We assign independent Bernoulli$(1/2)$ priors to $\gamma_1$ and $\gamma_2$, and independent $\mathcal{N}(0,100)$ priors to the regression coefficients. The target is the joint posterior distribution of $\boldsymbol{\gamma}$ and the corresponding active coefficient vector $\boldsymbol{\theta}_{\boldsymbol{\gamma}}$.

We also consider an eight-model extension in which an additional scalar block is appended to the same robust-regression design, giving an enlarged covariate vector $\mathbf{x}_i^{\mathrm{ex}}\in\mathbb{R}^5$ and coefficient vector $\boldsymbol{\beta}^{\mathrm{ex}}=(\beta_0,\ldots,\beta_4)$. In this case
$$
\boldsymbol{\gamma}^{\mathrm{ex}}=(1,\gamma_1,\gamma_2,\gamma_3)\in\{1\}\times\{0,1\}^3,
\qquad
\mathbf{m}_{\mathrm{ex}}(\boldsymbol{\gamma}^{\mathrm{ex}})
=(1,\gamma_1,\gamma_2,\gamma_2,\gamma_3),
$$
with block sizes $(1,1,2,1)$. The eight candidate models are indexed by the binary strings
$1000,1001,1010,1011,1100,1101,1110,$ and $1111$, and the active dimension is
$d_{\boldsymbol{\gamma}^{\mathrm{ex}}}=1+\gamma_1+2\gamma_2+\gamma_3$. This extension is useful for assessing the conditional transport proposal in a larger saturated space, where training separate model-specific flows is less attractive.

The simulated data set is generated under the sparse block model $\boldsymbol{\gamma}^\star=(1,1,0)$, so that only $\beta_0$ and $\beta_1$ are active. As in \citet{C_AISTATS_davies2023transport}, half of the observations are generated using $(\beta_0,\beta_1)=(1,1)$ and the remaining half using $(\beta_0,\beta_1)=(6,1)$; the non-intercept covariates are drawn independently from $\mathcal{N}(0,1)$ and the simulation errors from $\mathcal{N}(0,25)$. The two regression components have the same slope but well-separated intercepts, so the data cannot be summarized well by a single Gaussian regression surface. Under the robust mixture likelihood, this produces multiple high-probability regions within the model-specific posteriors, making the example a useful stress test for between-model proposals rather than only for within-model local exploration.

The formal comparison uses the standard auxiliary proposal of \citet{J_JRSSSB_brooks2003efficient}, affine-map TRJ, RQMA-map TRJ, and RQMA-map CTP as sample-trained baselines. The first two fit model-specific maps, whereas CTP fits one conditional map over the saturated space. We compare them with the individual-map VI-NFs TRJ proposal of Section~\ref{sec:trj-individual} and the target-trained TVI-CNFs CTP proposal of Section~\ref{sec:ctp-vi}.

For the final-sampler comparisons, we use stored SMC reference probabilities. We monitor the posterior probability of the top reference model: in the four-model problem this is model $111$, with $p_{\mathrm{ref}}=0.616453$; in the eight-model problem this is model $1110$, with $p_{\mathrm{ref}}=0.303619$. The first $10^4$ iterations are discarded as burn-in for running estimates and final summaries.

\begin{table}[!tbp]
\centering
\caption{Sampling metrics for the four-model robust-regression variable-selection example at $N=4000$. The wall-clock column is the recorded 100,000-iteration RJMCMC sampling phase in seconds (mean $\pm$ sample standard deviation across three runs). The reference value is $p_{\mathrm{ref}}=\Pr(\mathbf{m}=111\mid y)=0.616453$.}
\label{tab:vs-four-model-results}
\resizebox{\linewidth}{!}{%
\begin{tabular}{lccccc}
\toprule
Proposal & RJMCMC sampling (s) & $\hat p(\mathbf{m}=111)$ & $|\hat p-p_{\mathrm{ref}}|$ & Jump accept. & Switch \\
\midrule
Standard & $310.9\pm15.9$ & 0.6437 & 0.0272 & 0.0031 & 0.0023 \\
Affine TRJ & $467.2\pm62.9$ & 0.6261 & 0.0096 & 0.1747 & 0.1312 \\
RQMA-TRJ & $1009.0\pm188.8$ & 0.6205 & 0.0040 & 0.3732 & 0.2805 \\
RQMA-CTP & $1036.0\pm2.1$ & 0.6227 & 0.0063 & 0.2906 & 0.2188 \\
VI-NFs TRJ & $788.2\pm3.2$ & 0.6047 & 0.0117 & 0.3853 & 0.2889 \\
TVI-CNFs CTP & $748.6\pm5.1$ & 0.6149 & 0.0016 & 0.3007 & 0.2249 \\
\bottomrule
\end{tabular}
}
\end{table}

\begin{figure}[!tbp]
  \begin{center}
    \centerline{\includegraphics[width=\columnwidth]{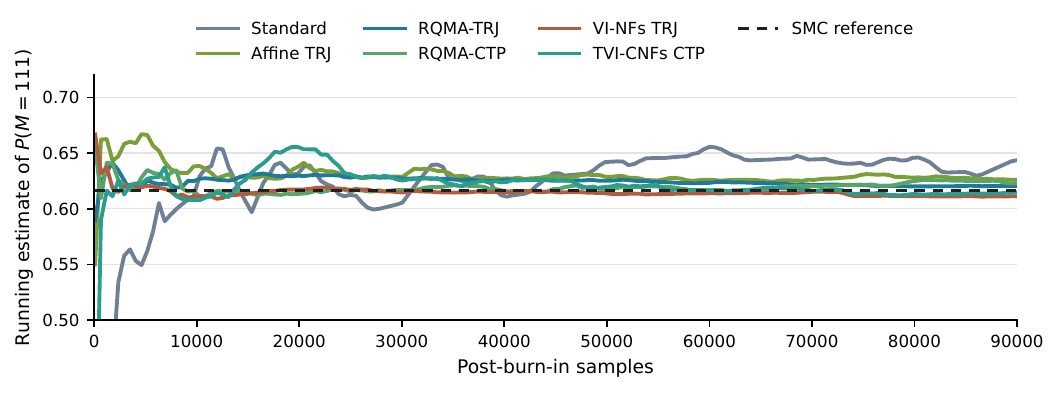}}
    \caption{Illustrative running-estimate diagnostic for the top reference model in the four-model robust-regression variable-selection example. The top reference model is $111$, with posterior probability $0.616453$.}
    \label{fig:vs-four-model-running}
  \end{center}
\end{figure}

\begin{table}[!tbp]
\centering
\caption{Sampling metrics for the eight-model CTP robust-regression variable-selection example. The wall-clock column is the recorded 200,000-iteration RJMCMC sampling phase in seconds (mean $\pm$ sample standard deviation across three runs). The reference value is $p_{\mathrm{ref}}=\Pr(\mathbf{m}=1110\mid y)=0.303619$.}
\label{tab:vs-eight-model-results}
\resizebox{\linewidth}{!}{%
\begin{tabular}{lccccc}
\toprule
Proposal & RJMCMC sampling (s) & $\hat p(\mathbf{m}=1110)$ & $|\hat p-p_{\mathrm{ref}}|$ & Jump accept. & Switch \\
\midrule
RQMA-CTP $N=1000$ & $2628.6\pm9.5$ & 0.3907 & 0.0871 & 0.3134 & 0.2742 \\
RQMA-CTP $N=2000$ & $2631.1\pm4.1$ & 0.5124 & 0.2088 & 0.2606 & 0.2278 \\
RQMA-CTP $N=4000$ & $2639.2\pm6.2$ & 0.3212 & 0.0176 & 0.3001 & 0.2620 \\
RQMA-CTP $N=8000$ & $2640.8\pm6.3$ & 0.2922 & 0.0114 & 0.4054 & 0.3554 \\
TVI-CNFs CTP $N=4000$ & $1639.8\pm37.6$ & 0.2872 & 0.0164 & 0.3813 & 0.3340 \\
\bottomrule
\end{tabular}
}
\end{table}

\begin{figure}[!tbp]
  \begin{center}
    \centerline{\includegraphics[width=\columnwidth]{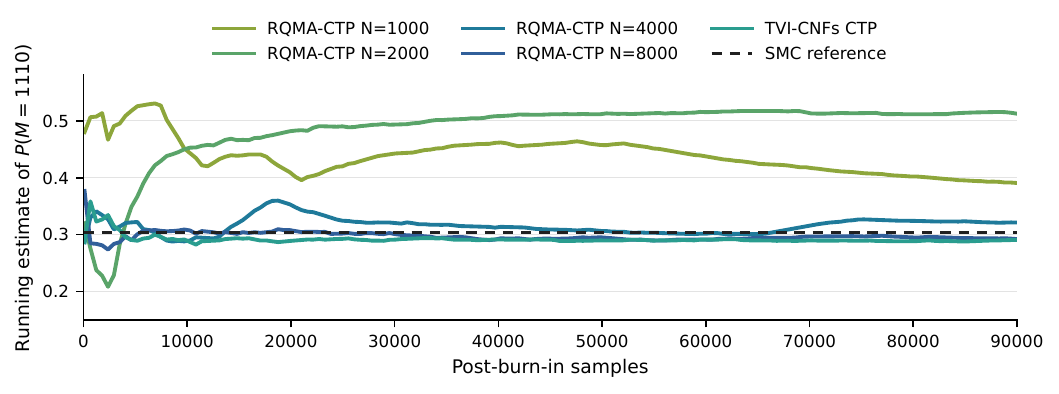}}
    \caption{Fixed qualitative running-estimate diagnostic for the posterior probability of the top model $1110$ in the eight-model robust-regression variable-selection example. The RQMA-CTP curves use particle-trained conditional transport proposals with $N=1000,2000,4000,$ and $8000$, and the TVI-CNFs CTP curve uses the reverse-KL-trained conditional normalizing-flow proposal. The dashed horizontal line denotes the reference probability $0.303619$.}
    \label{fig:vs-eight-model-running}
  \end{center}
\end{figure}

Table~\ref{tab:vs-four-model-results} shows that the formal fitted proposals recover the dominant four-model posterior probability while substantially increasing mobility relative to the standard saturated auxiliary proposal; Figure~\ref{fig:vs-four-model-running} provides the associated qualitative trajectories. In the eight-model CTP comparison, Table~\ref{tab:vs-eight-model-results} and Figure~\ref{fig:vs-eight-model-running} show that the conditional VI proposal gives a competitive top-model estimate using one conditional flow in the saturated space. At the setting, its recorded sampling phase is shorter than that of RQMA-CTP. Related pilot-sampling and map-fitting time records are reported in Supplementary Appendix~D.

\FloatBarrier

\subsection{Coal-mining disaster change-point example}
\label{sec:coal_mining}

We next consider the classical coal-mining disaster data, a standard benchmark for trans-dimensional Bayesian computation \citep{J_BIOMET_green1995reversible,BS_green2003transdimensional,J_SAC_fearnhead2006exact}. The data consist of the times $y=\{t_1,\ldots,t_n\}\subset[0,L]$ at which coal-mining disasters occurred. Under a model with $k$ change points, the observation window is divided into $k+1$ intervals, within each of which the events are modelled as a homogeneous Poisson process. Let $0=\tau_0<\tau_1<\cdots<\tau_k<\tau_{k+1}=L$ denote the segment boundaries, and let $\ell_i=\tau_i-\tau_{i-1}, N_i=\#\{t_j:\tau_{i-1}<t_j\le \tau_i\}, i=1,\ldots,k+1,$ be the corresponding segment lengths and event counts. Writing $h_i>0$ for the Poisson intensity in segment $i$, the likelihood is $p(y\mid k,\tau,h) \propto \prod_{i=1}^{k+1} h_i^{N_i}\exp(-\ell_i h_i), $ where $h=(h_1,\ldots,h_{k+1})$.

Following the prior specification used in the coal-mining example of \citet{J_BIOMET_green1995reversible}, we assign a truncated Poisson prior to the number of change points, $p(k)=\frac{\mathrm{Poisson}(k;3)}{\sum_{j=1}^{k_{\max}}\mathrm{Poisson}(j;3)}$, where $k=1,\ldots,k_{\max},$ and set $k_{\max}=6$. Conditional on $k$, we work with the normalized segment lengths $w_i=\frac{\ell_i}{L},w=(w_1,\ldots,w_{k+1})\in\Delta_k,$ where $\Delta_k=\{w_i>0,\sum_{i=1}^{k+1}w_i=1\}$. The prior induced by Green's construction, in which the change points are the even-numbered order statistics of $2k+1$ uniform random variables, is $w\mid k\sim \mathrm{Dirichlet}(2,\ldots,2).$ For the segment intensities, we use independent Gamma priors $ h_i\mid k \stackrel{\mathrm{i.i.d.}}{\sim} \Gamma(\alpha,\beta),\alpha=1, \beta=200,$ under the shape--rate parameterization. The nuisance intensities $h$ can be integrated out analytically. For fixed $(k,w)$, the collapsed likelihood is $p(y\mid k,w) \propto \prod_{i=1}^{k+1} \frac{\Gamma(\alpha+N_i)}{\Gamma(\alpha)} \frac{\beta^\alpha}{(\beta+\ell_i)^{\alpha+N_i}}, \ell_i=Lw_i$. Thus the collapsed posterior targeted by our reversible-jump algorithms is $ \pi(k,w\mid y) \propto p(k)\,p(w\mid k)\,p(y\mid k,w), k=1,\ldots,k_{\max}$.

For this example, we use the posterior model probabilities reported by \citet{BS_green2003transdimensional} as reference values. As discussed by \citet{J_SAC_fearnhead2006exact}, the original RJMCMC run of \citet{J_BIOMET_green1995reversible} had not fully converged for this data set, whereas the reanalysis of \citet{BS_green2003transdimensional}, using the same model and RJMCMC algorithm but run for 25 times as many iterations, gave results that were almost identical to those obtained by exact simulation. The resulting reference posterior probabilities for the number of change points are
\begin{equation}
\begin{array}{c|cccccc}
k & 1 & 2 & 3 & 4 & 5 & 6\\
\hline
\pi_{\mathrm{ref}}(k\mid y)
& 0.058 & 0.251 & 0.294 & 0.236 & 0.117 & 0.044
\end{array}
\end{equation}
which we use as benchmark reference values for evaluating posterior model probability estimates.

\begin{table}[!tbp]
\centering
\caption{Sampling and model-probability diagnostics for the coal-mining change-point example.  The wall-clock column is the recorded 100,000-iteration RJMCMC sampling phase in seconds (mean $\pm$ sample standard deviation across three runs).  The reference value is $p_{\mathrm{ref}}=\Pr(K=3\mid y)=0.294$.}
\label{tab:cp-total-time}
\resizebox{\linewidth}{!}{%
\begin{tabular}{lccccc}
\toprule
Proposal & RJMCMC sampling (s) & $\hat p(K=3)$ & $|\hat p-p_{\mathrm{ref}}|$ & Jump accept. & Switch \\
\midrule
RQMA-CTP $N=1000$ & $1182.2\pm6.4$ & 0.3045 & 0.0105 & 0.0277 & 0.0139 \\
RQMA-CTP $N=2000$ & $1140.2\pm82.1$ & 0.3286 & 0.0346 & 0.0316 & 0.0158 \\
RQMA-CTP $N=4000$ & $1186.6\pm6.5$ & 0.2923 & 0.0017 & 0.0276 & 0.0138 \\
RQMA-CTP $N=8000$ & $1179.8\pm21.9$ & 0.3246 & 0.0306 & 0.0279 & 0.0140 \\
TVI-CNFs CTP  & $772.3\pm21.4$ & 0.3074 & 0.0134 & 0.0490 & 0.0245 \\
\bottomrule
\end{tabular}
}
\end{table}

Table~\ref{tab:cp-total-time} shows that the RQMA-CTP estimates vary with particle-calibration size, with $N=4000$ giving the closest retained estimate of the reference posterior probability $P(K=3)=0.294$.  At the matched setting, TVI-CNFs CTP remains within about one percentage point of the reference and has the largest retained jump-acceptance and switching rates, indicating more frequent accepted movement across change-point models. Related pilot-sampling and map-fitting time records are reported in Supplementary Appendix D. Figure~\ref{fig:cp-running-k3} gives the corresponding qualitative running estimates.

\begin{figure}[!tbp]
  \begin{center}
    \centerline{\includegraphics[width=\columnwidth]{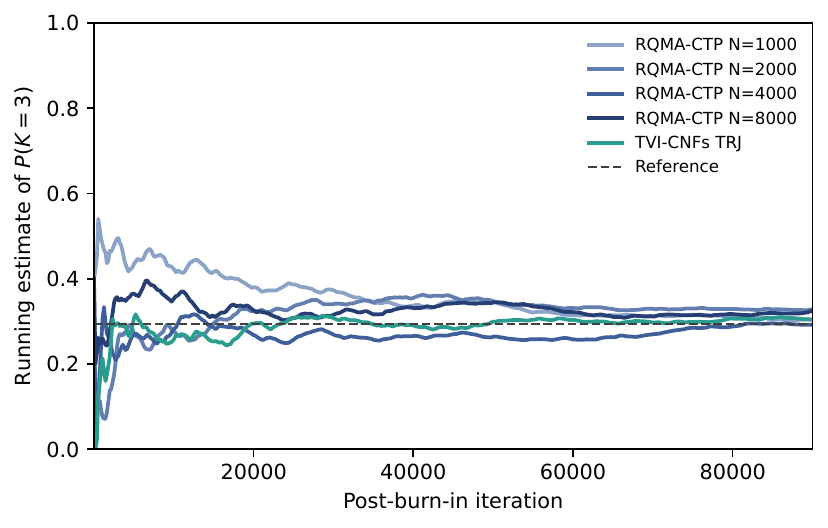}}
    \caption{Fixed qualitative running-estimate diagnostic for the posterior probability $P(K=3)$ in the change-point model-selection example.  The RQMA-CTP curves correspond to transport proposals calibrated from training particles, with training sample sizes $N=1000,2000,4000,$ and $8000$.  The TVI-CNFs CTP curve corresponds to the reverse-KL-trained conditional normalizing-flow proposal.  Each curve shows the average running estimate over independent RJMCMC chains after discarding the first $10^4$ iterations as burn-in.  The dashed horizontal line denotes the reference posterior probability $P(K=3)=0.294$.}
    \label{fig:cp-running-k3}
  \end{center}
\end{figure}

\subsection{Phylogenetic substitution-model example}
\label{sec:phylo_substitution}

We finally consider Bayesian phylogenetic substitution-model selection, following the reversible-jump framework of \citet{J_MBAE_huelsenbeck2004bayesian}. The observations are DNA sequence alignments, denoted by $\mathbf{x}$. At site $\ell$, taxon $i$ has nucleotide $x_{i\ell}\in\{A,C,G,T\}$. We use two data sets from \citet{J_MBAE_huelsenbeck2004bayesian}: the Primates alignment, consisting of $12$ taxa and $898$ sites from partial mitochondrial DNA, and the Vertebrates alignment, consisting of $17$ taxa and $432$ sites from the $\beta$-globin gene. In the original analysis of \citet{J_MBAE_huelsenbeck2004bayesian}, the unrooted tree topology is included in the Markov-chain state. Here, to isolate the trans-dimensional substitution-model component, the tree topology is fixed at $s_0$, and posterior inference is performed over the substitution model and the remaining continuous parameters, including branch lengths, nucleotide frequencies, the among-site rate-variation parameter, and substitution-rate parameters.

Let $\mathcal{A}=\{A,C,G,T\}$ denote the nucleotide alphabet and  the six unordered nucleotide substitution types denoted as $\mathcal{E}=\{AC,AG,AT,CG,CT,GT\}$. A substitution model $m\in\mathcal{M}$ is a partition of $\mathcal{E}$. We write $\mathcal{G}_{m,g}$ for group $g$ under model $m$, and $n_{m,g}=|\mathcal{G}_{m,g}|$ for the size of that group. The number of distinct substitution-rate groups is denoted by $K_m$. Since the six substitution types can be partitioned in $\mathrm{Bell}(6)=203$ ways, the model space has $|\mathcal{M}|=203$ elements. Equivalently, each model can be represented by a six-digit string in the order $\mathcal{E}$, where equal digits impose equality constraints on the corresponding substitution rates. For example, $111111$ is the most restrictive model with a single substitution rate, whereas $123456$ is the general time-reversible model.

For model $m$, let $\boldsymbol{\rho}_m=(\rho_{m,1},\ldots,\rho_{m,K_m})$ denote the distinct substitution-rate parameters. The exchangeability $r_{XY}$ for $XY\in\mathcal{E}$ is defined by $r_{XY}=\rho_{m,g}$ whenever $XY\in\mathcal{G}_{m,g}$. Let $\boldsymbol{\pi}=(\pi_A,\pi_C,\pi_G,\pi_T)\in\Delta_3$ be the nucleotide-frequency vector. Conditional on model $m$, the time-reversible rate matrix $\mathbf{Q}_m(\boldsymbol{\rho}_m,\boldsymbol{\pi})$ has off-diagonal entries $q_{XY}=r_{XY}\pi_Y$ for $X\neq Y$, and diagonal entries chosen so that each row sums to zero. The matrix is then rescaled so that branch lengths are measured in expected substitutions per site, equivalently by normalizing the mean substitution rate to one. Let $\mathbf{v}=(v_1,\ldots,v_B)$ denote the branch lengths on the fixed topology $s_0$, where $B=2n_{\mathrm{tax}}-3$ for a fully resolved unrooted tree, and let $\alpha>0$ denote the gamma shape parameter controlling among-site rate variation. The model-specific continuous parameter vector is $\boldsymbol{\theta}_m=(\mathbf{v},\alpha,\boldsymbol{\pi},\boldsymbol{\rho}_m)$. Since $\boldsymbol{\pi}$ has three free components and $\boldsymbol{\rho}_m$ has $K_m-1$ free components after the normalization constraint, the dimension of the continuous parameter space is $d_m=B+1+3+(K_m-1)=B+K_m+3$. This example is therefore trans-dimensional because both the model index $m$ and the dimension of $\boldsymbol{\theta}_m$ change across models.

Following \citet{J_MBAE_huelsenbeck2004bayesian}, we assign a uniform prior over the $203$ substitution models, $\pi(m)=1/203$ for $m\in\mathcal{M}$. Conditional on $m$, the branch lengths are assigned independent exponential priors $v_b\stackrel{\mathrm{i.i.d.}}{\sim}\mathrm{Exp}(10)$ for $b=1,\ldots,B$, the nucleotide frequencies have the flat Dirichlet prior $\boldsymbol{\pi}\sim\mathrm{Dirichlet}(1,1,1,1)$, and the gamma shape parameter has prior $\alpha\sim\mathrm{Uniform}(0,50)$. For the substitution-rate parameters, we first draw $\mathbf{u}_m=(u_{m,1},\ldots,u_{m,K_m})\sim\mathrm{Dirichlet}_{K_m}(1,\ldots,1)$ and then set $\rho_{m,g}=6u_{m,g}/n_{m,g}$ for $g=1,\ldots,K_m$. Equivalently, the prior on $\boldsymbol{\rho}_m$ is supported on positive rates satisfying $\sum_{g=1}^{K_m}n_{m,g}\rho_{m,g}=6$, so that the six exchangeabilities have average one before the final rate-matrix scaling. When $K_m=1$, this construction gives the single-rate model $\rho_{m,1}=1$.

\begin{sloppypar}
We assume sites are conditionally independent given the phylogenetic parameters. Let $\omega_c(\alpha)$, $c=1,\ldots,4$, denote the four discrete-gamma rate multipliers used to approximate among-site rate variation, normalised to have mean one. For a rate matrix $\mathbf{Q}$, let $\mathcal{L}_{\ell}(s_0,\mathbf{v},\mathbf{Q})$ denote the site likelihood at site $\ell$, computed on the fixed topology $s_0$ by the pruning recursion of \citet{J_JME_felsenstein1981evolutionary}, with branch transition matrices $\mathbf{P}_b=\exp(v_b\mathbf{Q})$ for $b=1,\ldots,B$. To avoid cumbersome notation, let $\bar{\mathcal{L}}_{\ell}(m,\boldsymbol{\theta}_m)$ denote the corresponding site likelihood averaged over the four discrete-gamma rate categories. The likelihood under model $m$ is then $p(\mathbf{x}\mid m,\boldsymbol{\theta}_m,s_0)=\prod_{\ell=1}^{L}\bar{\mathcal{L}}_{\ell}(m,\boldsymbol{\theta}_m)$. The resulting fixed-topology posterior distribution is $\pi(m,\boldsymbol{\theta}_m\mid\mathbf{x},s_0)\propto p(\mathbf{x}\mid m,\boldsymbol{\theta}_m,s_0)\pi(\boldsymbol{\theta}_m\mid m)\pi(m)$ for $m\in\mathcal{M}$, where $\pi(\boldsymbol{\theta}_m\mid m)$ collects the priors on branch lengths, nucleotide frequencies, the gamma shape parameter, and substitution-rate parameters. Compared with the original posterior of \citet{J_MBAE_huelsenbeck2004bayesian}, the topology prior and topology updates are absent because $s_0$ is fixed throughout our experiments.
\end{sloppypar}

To sample from this fixed-topology posterior, we compare two reversible-jump algorithms. The first is the fixed-topology version of the split--merge RJMCMC sampler of \citet{J_MBAE_huelsenbeck2004bayesian}, which we use as the baseline. Its trans-dimensional moves propose changes in $m$ by either merging two existing substitution groups or splitting one substitution group into two groups, while the remaining continuous parameters are updated within model. The second is RJMCMC with our CTP proposal. For CTP, the constrained parameters are first represented in unconstrained transport coordinates, and a single conditional proposal is trained across the $203$ substitution models using the conditional Gaussian base and conditional normalizing-flow construction of Section~\ref{sec:ctp-vi}. Both algorithms use the same fixed tree topology $s_0$, the same likelihood, and the same prior specification.

For each of the Primates and Vertebrates alignments, we monitor both the posterior distribution over the $203$ substitution models and the induced posterior distribution of the number of substitution types, $\pi(K=k\mid\mathbf{x},s_0) = \sum_{m\in\mathcal{M}:K_m=k} \pi(m\mid\mathbf{x},s_0), k=1,\ldots,6$. All comparisons are made under the fixed-topology posterior with tree topology $s_0$. To provide a stable reference for the running estimates, we additionally ran three independent 1M-iteration standard chains and use their average posterior model probabilities as the reference values; these long chains are used only for reference and are not included in the 200k efficiency comparison.

\begin{table}[!tbp]
\centering
\caption{Posterior model-probability and sampling diagnostics for the phylogenetic substitution-model examples. Results are empirical means over three independently seeded 200k-iteration MCMC chains for each method and data set, with the first 20\% discarded as burn-in. The column $\hat p$ gives the posterior probability estimate of the displayed top model, and Ref. gives the corresponding estimate from three independent 1M standard chains.}
\label{tab:phylo-results}
\resizebox{\linewidth}{!}{%
\begin{tabular}{lllccccc}
\toprule
Dataset & Method & Top model & $\hat p$ & Ref. & $|\hat p-\mathrm{Ref.}|$ & Jump accept. & Switch \\
\midrule
Primates & Standard & \texttt{121121} & 0.2166 & 0.2457 & 0.0291 & 0.0971 & 0.0485 \\
Primates & TVI-CNFs CTP & \texttt{121121} & 0.2447 & 0.2457 & 0.0087 & 0.0966 & 0.0504 \\
Vertebrates & Standard & \texttt{123343} & 0.2186 & 0.1800 & 0.0386 & 0.0842 & 0.0421 \\
Vertebrates & TVI-CNFs CTP & \texttt{123343} & 0.1798 & 0.1800 & 0.0002 & 0.1368 & 0.0684 \\
\bottomrule
\end{tabular}
}
\end{table}

Table~\ref{tab:phylo-results} show that both samplers identify the same dominant substitution model on each data set. For the Primates alignment, the TVI-CNFs CTP estimates the top-model probability as $0.2447$, compared with the long-run reference value $0.2457$ (an absolute difference of $0.0087$), with jump acceptance comparable to that of the standard split--merge proposal and a slightly higher model-switching rate. For the Vertebrates alignment, the conditional transport gives a top-model probability estimate close to the reference value and higher jump acceptance and model-switching rates. Thus, in this fixed-topology phylogenetic example, the  TVI-CNFs CTP preserves the qualitative model-selection conclusion of the baseline sampler and gives the clearest mobility gain on the Vertebrates data set.

As a qualitative comparison with the original MBE2004 analysis, we also examine the highest-posterior models and the reported 95\% credible sets in Table~3 of \citet{J_MBAE_huelsenbeck2004bayesian}. Although that analysis marginalizes over tree topologies whereas ours conditions on $s_0$, the dominant selected models agree. For the Primates data, both standard and TVI-CNFs CTP select M15 (\texttt{121121}), the same highest-posterior model reported in MBE2004, and their estimated 95\% credible sets contain all 15 models in the MBE2004 credible set. For the Vertebrates $\beta$-globin data, both methods select M125 (\texttt{123343}), again matching the MBE2004 highest-posterior model. The standard chain contains 16 of the 17 MBE2004 credible-set models, missing only the boundary model \texttt{112342}, while TVI-CNFs CTP contains 15 of the 17 models, missing \texttt{112234} and \texttt{112342}. These discrepancies occur only at the low-posterior boundary of the 95\% credible set; the dominant posterior model and the main transition--transversion rate structure are unchanged across MBE2004, standard, and TVI-CNFs CTP.

\FloatBarrier
\section{Discussion}
\label{sec:discussion}

We have developed a posterior-exact framework for training transport-map proposals for RJMCMC by variational inference with normalizing flows.  The construction applies to both between-model and within-model proposals and, unlike sample-trained transports, fits the maps using reference draws and evaluations of unnormalized targets rather than pilot draws from difficult conditional posteriors.  The fitted flows are used only to construct proposals: the Metropolis--Hastings correction retains the original trans-dimensional posterior as the invariant distribution.

Our use of RealNVP is motivated by the computational requirements of transport RJMCMC.  Its coupling structure permits parallel evaluation within each transformation while providing the forward map, inverse map, and Jacobian terms required repeatedly in trans-dimensional proposals.  Combined with the saturated-space conditional construction, this yields a single variational training problem that shares information across model dimensions rather than fitting a separate map from target samples for every model.

The numerical results show that reverse-KL-trained transports can give accurate posterior model-probability estimates and effective trans-dimensional exploration without a pilot posterior-simulation stage for map fitting.  Across the controlled and applied examples, the conditional flow construction is competitive with, and in several settings improves on, sample-trained CTP proposals in model-probability accuracy and switching efficiency.  The gains are naturally problem dependent: imperfect reverse-KL maps may under-cover parts of a conditional posterior and thereby affect finite-time acceptance and mixing, but they do not introduce asymptotic bias after the RJMCMC correction.

The framework also suggests several extensions.  The variational maps provide importance sampling estimates of model evidences as a by-product; when these are sufficiently accurate, they can be used to construct model-index proposals close to posterior model probabilities and hence approach rejection-free trans-dimensional moves.  It is also natural to extend the conditional construction to variational approximations of the joint posterior $\pi(k,\boldsymbol{\theta}_k\mid\mathbf y)$ by introducing an explicit variational distribution for the model index.

\section*{Supplementary material}
The supplementary material is included after the references. Appendix A gives proofs for the section \ref{sec:approx_rkl_theory_revised}. Appendix B gives the individual and conditional stochastic-gradient fitting routines. Appendix C gives flow-depth ablations. Appendix D reports pilot-sampling, sample-trained map-fitting, reverse-KL flow-training, and phylogenetic sampling times.

\bibliographystyle{ba}
\bibliography{bibs/abbr}

\clearpage
\phantomsection
\pdfbookmark[0]{Supplementary Material}{supplementary-material}
\input{appendices}

\end{document}

%% file: appendices.tex
\section*{Appendix A: Proofs and Additional Calculations for Section 3}
\label{app:proofs_sec35}

Throughout this appendix, we use the notation introduced in the main text. These results apply only to individual transports with the reference law $\otimes_{d_k}\nu$; they do not provide the corresponding simplification for conditional flows with context-dependent bases. We interpret $\otimes_0\nu$ as the unit density. For model $k\in\mathcal K$, write $\pi_k(\theta)$ for the conditional target density on $\Theta_k$ and $d_k=\dim(\Theta_k)$. For a diffeomorphism $T_k:\Theta_k\to\mathbb R^{d_k}$, define
\[
q_k(\theta)
:=
(\otimes_{d_k}\nu)\{T_k(\theta)\}\,\lvert\det\nabla T_k(\theta)\rvert,
\qquad
w_k(\theta):=\frac{\pi_k(\theta)}{q_k(\theta)}.
\]
The density of the pushforward $T_k\#\pi_k$ on reference space is denoted by
\[
r_k(z)
:=
\pi_k\{T_k^{-1}(z)\}\,\lvert\det\nabla T_k^{-1}(z)\rvert .
\]
The model-index proposal probability from $k$ to $k'$ is denoted by $q(k'\mid k)$.

For a between-model move with $d_{k'}\ge d_k$, draw $u\sim\otimes_{d_{k'}-d_k}\nu$, set
\[
z=T_k(\theta),
\qquad
z'=\bar h_{k,k'}(z,u),
\qquad
\theta'=T_{k'}^{-1}(z').
\]
The reference-space map $\bar h_{k,k'}$ is assumed to be volume-preserving and reference-invariant, so that
\[
(\otimes_{d_{k'}}\nu)\{\bar h_{k,k'}(z,u)\}
=
(\otimes_{d_k}\nu)(z)\,(\otimes_{d_{k'}-d_k}\nu)(u).
\]
The case $d_{k'}<d_k$ is defined analogously using the inverse dimension-matching map.

\subsection*{A.1 Proof of Proposition 3.1}

We prove the case $d_{k'}\ge d_k$; the reverse-dimension case follows by applying the same argument to the inverse dimension-matching map. The RJMCMC ratio for the transport-based proposal is
\[
R_T
=
\frac{
\pi(k'\mid\mathbf y)\pi_{k'}(\theta')q(k\mid k')\lvert\det\nabla T_k(\theta)\rvert
}{
\pi(k\mid\mathbf y)\pi_k(\theta)q(k'\mid k)(\otimes_{d_{k'}-d_k}\nu)(u)\lvert\det\nabla T_{k'}(\theta')\rvert
}.
\]
By reference invariance,
\[
(\otimes_{d_{k'}}\nu)(z')
=
(\otimes_{d_k}\nu)(z)(\otimes_{d_{k'}-d_k}\nu)(u).
\]
Using the definition
\[
q_j(\vartheta)=(\otimes_{d_j}\nu)\{T_j(\vartheta)\}\lvert\det\nabla T_j(\vartheta)\rvert,
\]
we obtain
\[
\frac{\lvert\det\nabla T_k(\theta)\rvert}
{(\otimes_{d_{k'}-d_k}\nu)(u)\lvert\det\nabla T_{k'}(\theta')\rvert}
=
\frac{q_k(\theta)}{q_{k'}(\theta')}.
\]
Substitution into the RJMCMC ratio gives
\[
R_T
=
\frac{\pi(k'\mid\mathbf y)q(k\mid k')}
{\pi(k\mid\mathbf y)q(k'\mid k)}
\frac{\pi_{k'}(\theta')/q_{k'}(\theta')}{\pi_k(\theta)/q_k(\theta)}
=
R_\star(k,k')\frac{w_{k'}(\theta')}{w_k(\theta)},
\]
where
\[
R_\star(k,k'):=
\frac{\pi(k'\mid\mathbf y)q(k\mid k')}
{\pi(k\mid\mathbf y)q(k'\mid k)}.
\]
This is the factorization asserted in Proposition 3.1. If the maps are exact, then $T_k\#\pi_k=\otimes_{d_k}\nu$ and the change-of-variables formula gives
\[
\pi_k(\theta)=(\otimes_{d_k}\nu)\{T_k(\theta)\}\lvert\det\nabla T_k(\theta)\rvert=q_k(\theta).
\]
Thus $w_k\equiv1$ for all $k$, and $R_T=R_\star(k,k')$.

\subsection*{A.2 Proof of Corollary 3.2}

By Proposition 3.1,
\[
R_T
=
R_\star(k,k')\exp\{\Delta_{k,k'}(\theta,\theta')\},
\qquad
\Delta_{k,k'}(\theta,\theta')
:=
\log w_{k'}(\theta')-\log w_k(\theta).
\]
Therefore
\[
\alpha_T
:=1\wedge R_T
=
1\wedge\{R_\star(k,k')\exp(\Delta_{k,k'}(\theta,\theta'))\}.
\]
If $|\Delta_{k,k'}(\theta,\theta')|\le\eta$, then
\[
1\wedge(R_\star e^{-\eta})
\le
\alpha_T
\le
1\wedge(R_\star e^{\eta}).
\]
For any $a>0$ and $c\in(0,1]$, $1\wedge(ca)\ge c(1\wedge a)$. Taking $c=e^{-\eta}$ gives the lower bound. The upper bound follows from $1\wedge(e^{\eta}a)\le e^{\eta}(1\wedge a)$. Hence, with $\alpha_\star:=1\wedge R_\star(k,k')$,
\[
e^{-\eta}\alpha_\star
\le
\alpha_T
\le
e^{\eta}\alpha_\star.
\]

\subsection*{A.3 Conductance comparison for between-model moves}

Let \(B\subset\mathcal K\) and define
\[
A_B
:=
\bigcup_{k\in B}\{k\}\times\Theta_k .
\]
For a current state \(\omega=(k,\theta_k)\in A_B\) and a proposed model
\(k'\in B^c\), let \(\Lambda_{k,k'}\) denote the distribution of the
auxiliary random variables used in the dimension-matching proposal. For
example, if \(d_{k'}\ge d_k\), then
\(\Lambda_{k,k'}=\otimes_{d_{k'}-d_k}\nu\). If no auxiliary variable is
generated in the forward construction, \(\Lambda_{k,k'}\) is interpreted as
the unit mass on a singleton.

Define the common proposal-attempt measure
\[
\mathcal M_B(d\omega,dk',du)
=
\mathbf 1\{\omega\in A_B,\ k'\in B^c\}
\Pi(d\omega)\,
q(k'\mid k)\,
\Lambda_{k,k'}(du),
\]
where \(\Pi\) is the joint target distribution on the trans-dimensional
state space and \(q(k'\mid k)\) is the model-index proposal probability. This
measure contains the stationary law of the current state, the probability
of attempting a move to model \(k'\), and the law of the auxiliary
variables.

For the learned-map proposal, the stationary cross-cut flow is
\[
Q_T(A_B,A_B^c)
=
\int
\alpha_T(\omega,k',u)\,
\mathcal M_B(d\omega,dk',du),
\]
where \(\alpha_T\) is the Metropolis--Hastings acceptance probability using
the learned transport maps. Indeed, an accepted proposal with
\(k'\in B^c\) necessarily lands in \(A_B^c\), whereas a rejected proposal
remains in \(A_B\).

The exact-map idealized flow is
\[
Q_\star(A_B,A_B^c)
=
\int
\alpha_\star(k,k')\,
\mathcal M_B(d\omega,dk',du),
\]
where
\[
\alpha_\star(k,k')
=
1\wedge R_\star(k,k')
\]
is the acceptance probability in the idealized case where the transport
maps are exact. The learned-map and exact-map proposals need not induce the
same distribution on the proposed parameter value, but for the model-cut
event \(A_B\to A_B^c\) they are integrated with respect to the same
proposal-attempt measure \(\mathcal M_B\); the comparison is through the
acceptance probabilities.

Under the assumed uniform bound
\[
\left|
\Delta_{k,k'}(\theta_k,\theta_{k'}')
\right|
\le
\eta_B
\]
for attempted cross-cut moves, except on a
\(\mathcal M_B\)-null set, Corollary 3.2 gives the pointwise inequality
\[
\alpha_T(\omega,k',u)
\ge
e^{-\eta_B}
\alpha_\star(k,k')
\qquad
\mathcal M_B\text{-a.e.}
\]
Therefore,
\[
\begin{aligned}
Q_T(A_B,A_B^c)
&=
\int
\alpha_T(\omega,k',u)\,
\mathcal M_B(d\omega,dk',du) \\
&\ge
e^{-\eta_B}
\int
\alpha_\star(k,k')\,
\mathcal M_B(d\omega,dk',du) \\
&=
e^{-\eta_B}
Q_\star(A_B,A_B^c).
\end{aligned}
\]

Finally, the corresponding model-cut conductance is obtained by normalizing
the cross-cut flow by the stationary mass \(\Pi(A_B)\). Since the learned-map
kernel and the exact-map idealized kernel have the same target distribution
\(\Pi\), this stationary mass is the same for both kernels. Hence, whenever
\(\Pi(A_B)>0\),
\[
\mathcal C_T(B)
=
\frac{Q_T(A_B,A_B^c)}{\Pi(A_B)}
\ge
e^{-\eta_B}
\frac{Q_\star(A_B,A_B^c)}{\Pi(A_B)}
=
e^{-\eta_B}
\mathcal C_\star(B).
\]
This establishes the comparison.

\subsection*{A.4 Proposal under-coverage calculation}

Let
\[
E_{k,\varepsilon}:=
\{\theta\in\Theta_k:q_k(\theta)>\varepsilon\pi_k(\theta)\},
\qquad
S_{k,\varepsilon}:=\Theta_k\setminus E_{k,\varepsilon}.
\]
If $A\subseteq S_{k,\varepsilon}$, then $q_k(\theta)\le\varepsilon\pi_k(\theta)$ for every $\theta\in A$, up to null sets. Integrating over $A$ gives
\[
q_k(A)\le\varepsilon\pi_k(A).
\]
For independent proposal draws $\Theta_{k,1}^\dagger,\ldots,\Theta_{k,N}^\dagger\sim q_k$, the probability of missing $A$ is
\[
\Pr\{\Theta_{k,1}^\dagger\notin A,\ldots,\Theta_{k,N}^\dagger\notin A\}
=
(1-q_k(A))^N.
\]
The previous inequality and Bernoulli's inequality yield
\[
(1-q_k(A))^N
\ge
(1-\varepsilon\pi_k(A))^N
\ge
1-N\varepsilon\pi_k(A).
\]

For the stationary TRJ proposal from a source model $j$ to the target model $k$, the proposed reference variable in the target dimension is obtained from $z_j\sim r_j$ by adjoining or discarding reference coordinates and applying the reference-space dimension-matching map. The same operation sends $\otimes_{d_j}\nu$ to $\otimes_{d_k}\nu$ by reference invariance. Since total variation distance cannot increase under Markov kernels and measurable pushforwards,
\[
\|\mu_{j\to k}-q_k\|_{\mathrm{TV}}
\le
\|r_j-(\otimes_{d_j}\nu)\|_{\mathrm{TV}},
\]
where $\mu_{j\to k}$ denotes the distribution of a candidate in model $k$ before the Metropolis--Hastings acceptance step. The change-of-variables calculation in Appendix~A.6 gives
\[
\mathrm{KL}(\otimes_{d_j}\nu\|r_j)=\mathrm{KL}(q_j\|\pi_j),
\]
and Pinsker's inequality gives
\[
\|r_j-(\otimes_{d_j}\nu)\|_{\mathrm{TV}}
\le
\{\mathrm{KL}(q_j\|\pi_j)/2\}^{1/2}.
\]
Therefore, for $A\subseteq S_{k,\varepsilon}$,
\[
\mu_{j\to k}(A)
\le
q_k(A)+\|\mu_{j\to k}-q_k\|_{\mathrm{TV}}
\le
\varepsilon\pi_k(A)+\{\mathrm{KL}(q_j\|\pi_j)/2\}^{1/2}.
\]
This proves the under-coverage calculation.

\subsection*{A.5 Sticky-exit calculation}

If $\theta\in S_{k,\varepsilon}$, then $q_k(\theta)\le\varepsilon\pi_k(\theta)$ and hence
\[
w_k(\theta)=\frac{\pi_k(\theta)}{q_k(\theta)}\ge\frac1\varepsilon
\]
wherever the ratio is finite. If the proposed destination state satisfies $w_{k'}(\theta')\le M$, then the factorization in Proposition 3.1 gives
\[
R_T
=
R_\star(k,k')\frac{w_{k'}(\theta')}{w_k(\theta)}
\le
R_\star(k,k')M\varepsilon.
\]
Taking $1\wedge$ of both sides yields
\[
\alpha_T
\le
1\wedge\{R_\star(k,k')M\varepsilon\},
\]
which is the claimed sticky-exit bound.

\subsection*{A.6 Reference-space flattening identity}

By the change of variables $z=T_k(\theta)$,
\[
q_k\{T_k^{-1}(z)\}
=
(\otimes_{d_k}\nu)(z)\lvert\det\nabla T_k^{-1}(z)\rvert^{-1},
\qquad
q_k(\theta)d\theta=(\otimes_{d_k}\nu)(z)dz.
\]
Therefore
\[
\begin{aligned}
\mathrm{KL}(q_k\|\pi_k)
&=
\int q_k(\theta)\log\frac{q_k(\theta)}{\pi_k(\theta)}d\theta \\
&=
\int (\otimes_{d_k}\nu)(z)
\log\frac{(\otimes_{d_k}\nu)(z)\lvert\det\nabla T_k^{-1}(z)\rvert^{-1}}
{\pi_k\{T_k^{-1}(z)\}}dz \\
&=
\int (\otimes_{d_k}\nu)(z)
\log\frac{(\otimes_{d_k}\nu)(z)}
{\pi_k\{T_k^{-1}(z)\}\lvert\det\nabla T_k^{-1}(z)\rvert}dz.
\end{aligned}
\]
Since
\[
r_k(z)=\pi_k\{T_k^{-1}(z)\}\lvert\det\nabla T_k^{-1}(z)\rvert,
\]
the last display is exactly $\mathrm{KL}(\otimes_{d_k}\nu\|r_k)$. Pinsker's inequality then gives
\[
\|r_k-(\otimes_{d_k}\nu)\|_{\mathrm{TV}}
\le
\{\mathrm{KL}(\otimes_{d_k}\nu\|r_k)/2\}^{1/2}
=
\{\mathrm{KL}(q_k\|\pi_k)/2\}^{1/2}.
\]

\FloatBarrier
\section*{Appendix B: Flow-Fitting Algorithms}
\label{sec:algorithms}

Algorithm~\ref{alg:sgd_nf} gives the individual reverse-KL fitting routine.
Algorithm~\ref{alg:sgd_CVINF} gives the optional beta-conditioned tempered
conditional fitting routine. Without this option, one sets $\beta=1$, omits
$\beta$ from the context.
Both algorithms describe flow fitting only; the final samplers apply the exact
Metropolis--Hastings-corrected reversible-jump transitions given in the main text.

\begin{algorithm}[tb]
\caption{Reverse-KL Fitting of an Individual Transport Map}
\label{alg:sgd_nf}
\KwIn{Unnormalized conditional target $\pi(\cdot,\mathbf y\mid k)$,
reference distribution $\otimes_{d_k}\nu$, flow $f_{\boldsymbol\eta}$,
minibatch size $m$, and $M$ optimization iterations}
Initialize $\boldsymbol\eta$\;
\For{$t\gets1$ \KwTo $M$}{
  Draw $\boldsymbol z^1,\ldots,\boldsymbol z^m
  \stackrel{\mathrm{iid}}{\sim}\otimes_{d_k}\nu$\;
  Set $\boldsymbol z^i_{\boldsymbol\eta}
  \gets f_{\boldsymbol\eta}(\boldsymbol z^i)$ for $i=1,\ldots,m$\;
  Estimate $\widehat\ell(\boldsymbol\eta)\gets m^{-1}\sum_{i=1}^m
  \log\left\{
  q_{\boldsymbol\eta}(\boldsymbol z^i_{\boldsymbol\eta})\,/
  \pi(\boldsymbol z^i_{\boldsymbol\eta},\mathbf y\mid k)
  \right\}$\;
  Update $\boldsymbol\eta$ with a stochastic optimizer using
  $\nabla_{\boldsymbol\eta}\widehat\ell(\boldsymbol\eta)$\;
}
\Return{$T_k^{-1}=f_{\boldsymbol\eta}$ and $T_k=f_{\boldsymbol\eta}^{-1}$}\;
\end{algorithm}

\begin{algorithm}[tb]
\caption{Trans-dimensional VI Fitting of Conditional Transport Maps}
\label{alg:sgd_CVINF}
\KwIn{Model-context distribution $\rho$; unnormalized augmented targets
$\{\tilde{\gamma}_k\}_{k=1}^K$; reference densities $\{r_k\}_{k=1}^K$;
conditional base distribution $\tilde q_{0,\boldsymbol\phi}$;
conditional RealNVP $\tilde f_{\boldsymbol\eta}$; minibatch size $m$;
number of iterations $M$; and $\beta_{\min}\in(0,1]$.}

\KwOut{Conditional transport maps
$\{\tilde T_k^{-1},\tilde T_k\}_{k=1}^K$ at $\beta=1$.}

Initialize $\boldsymbol\eta$ and $\boldsymbol\phi$\;

\For{$t\gets 1$ \KwTo $M$}{
    Draw $k^1,\ldots,k^m\stackrel{\mathrm{iid}}{\sim}\rho$\;

    Draw $\beta^1,\ldots,\beta^m
    \stackrel{\mathrm{iid}}{\sim}\mathcal U[\beta_{\min},1]$\;

    Set $\boldsymbol c^i\gets
    \boldsymbol c_{k^i,\beta^i}
    =(\boldsymbol c_{k^i},\beta^i)$ for $i=1,\ldots,m$\;

    Draw $\tilde{\boldsymbol z}_0^i
    \sim\tilde q_{0,\boldsymbol\phi}(\cdot\mid\boldsymbol c^i)$
    for $i=1,\ldots,m$\;

    Set $\tilde{\boldsymbol z}_{\boldsymbol\eta}^i
    \gets\tilde f_{\boldsymbol\eta}
    (\tilde{\boldsymbol z}_0^i\mid\boldsymbol c^i)$
    for $i=1,\ldots,m$\;

    Estimate
    \[
    \widehat{\mathcal L}_{\mathrm{temp}}
    \gets
    \frac{1}{m}\sum_{i=1}^m
    \left[
    \log\tilde q_{\boldsymbol\eta,\boldsymbol\phi}
    (\tilde{\boldsymbol z}_{\boldsymbol\eta}^i\mid\boldsymbol c^i)
    -
    \log\tilde\gamma_{k^i,\beta^i}
    (\tilde{\boldsymbol z}_{\boldsymbol\eta}^i)
    \right];
    \]

    Update $(\boldsymbol\eta,\boldsymbol\phi)$ using a stochastic optimizer
    and $\nabla_{\boldsymbol\eta,\boldsymbol\phi}
    \widehat{\mathcal L}_{\mathrm{temp}}$\;
}

\For{$k\gets 1$ \KwTo $K$}{
    Set $\tilde T_k^{-1}(\cdot)
    \gets\tilde f_{\boldsymbol\eta}
    (\cdot\mid\boldsymbol c_{k,1})$\;

    Set $\tilde T_k\gets(\tilde T_k^{-1})^{-1}$\;
}
\end{algorithm}

\FloatBarrier
\section*{Appendix C: Flow-depth Ablations}

These sensitivity analyses vary the number of RealNVP coupling layers while holding the remaining stored configuration fixed.  Each coloured curve is a running estimate of the posterior probability of the displayed reference model, and the horizontal black line is its reference probability.

\begin{figure}[!htbp]
  \centering
  \includegraphics[width=\linewidth]{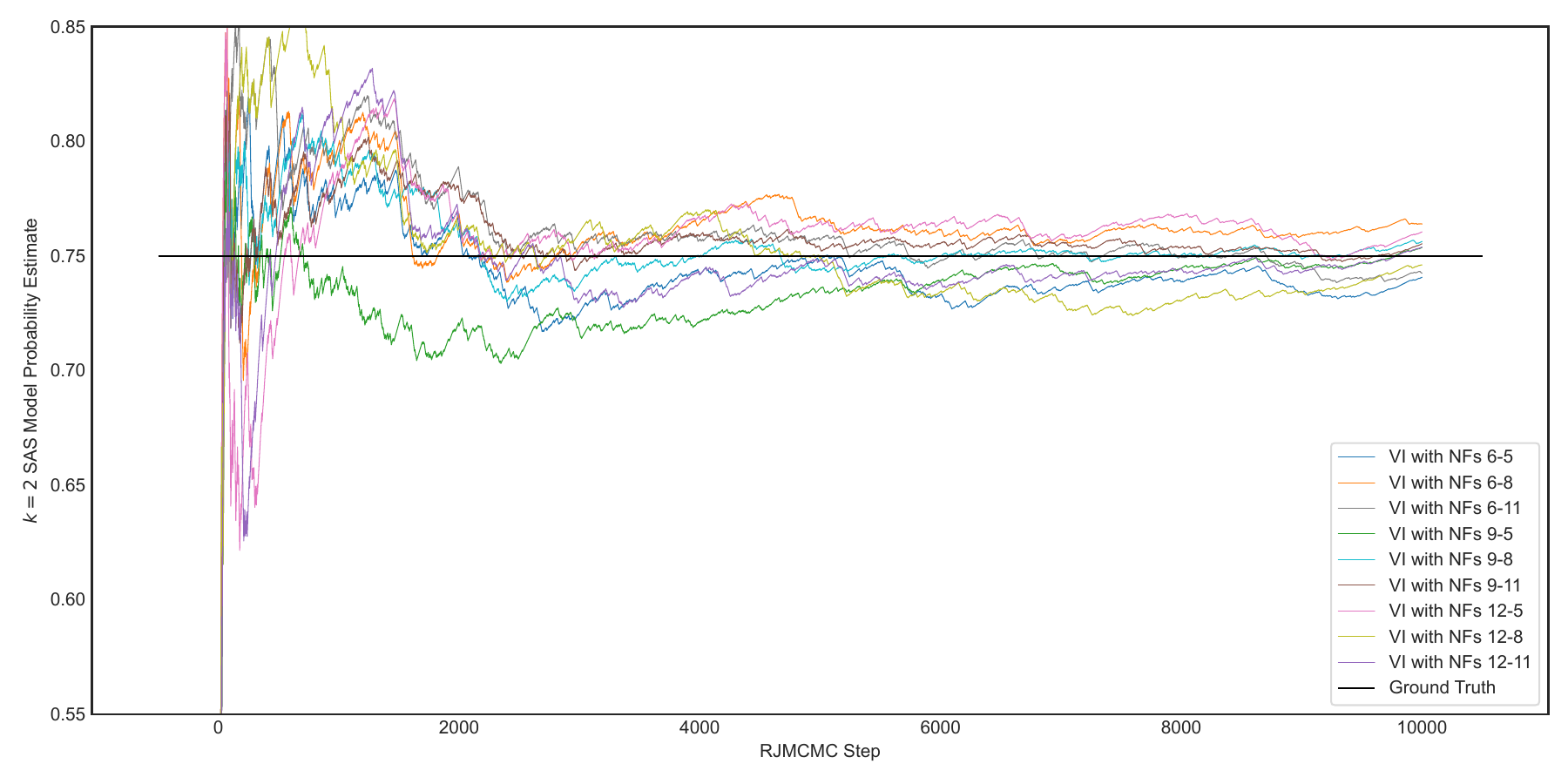}
  \caption{Flow-depth ablation for the illustrative SAS example.  The running model-probability traces compare the reverse-KL TRJ proposal under several flow-depth choices.  The plot is retained as a sensitivity diagnostic for the controlled example; the oracle-map comparison is given in the main paper.}
  \label{fig:SAS-ablation-depth}
\end{figure}

\begin{figure}[!htbp]
  \centering
  \includegraphics[width=\linewidth]{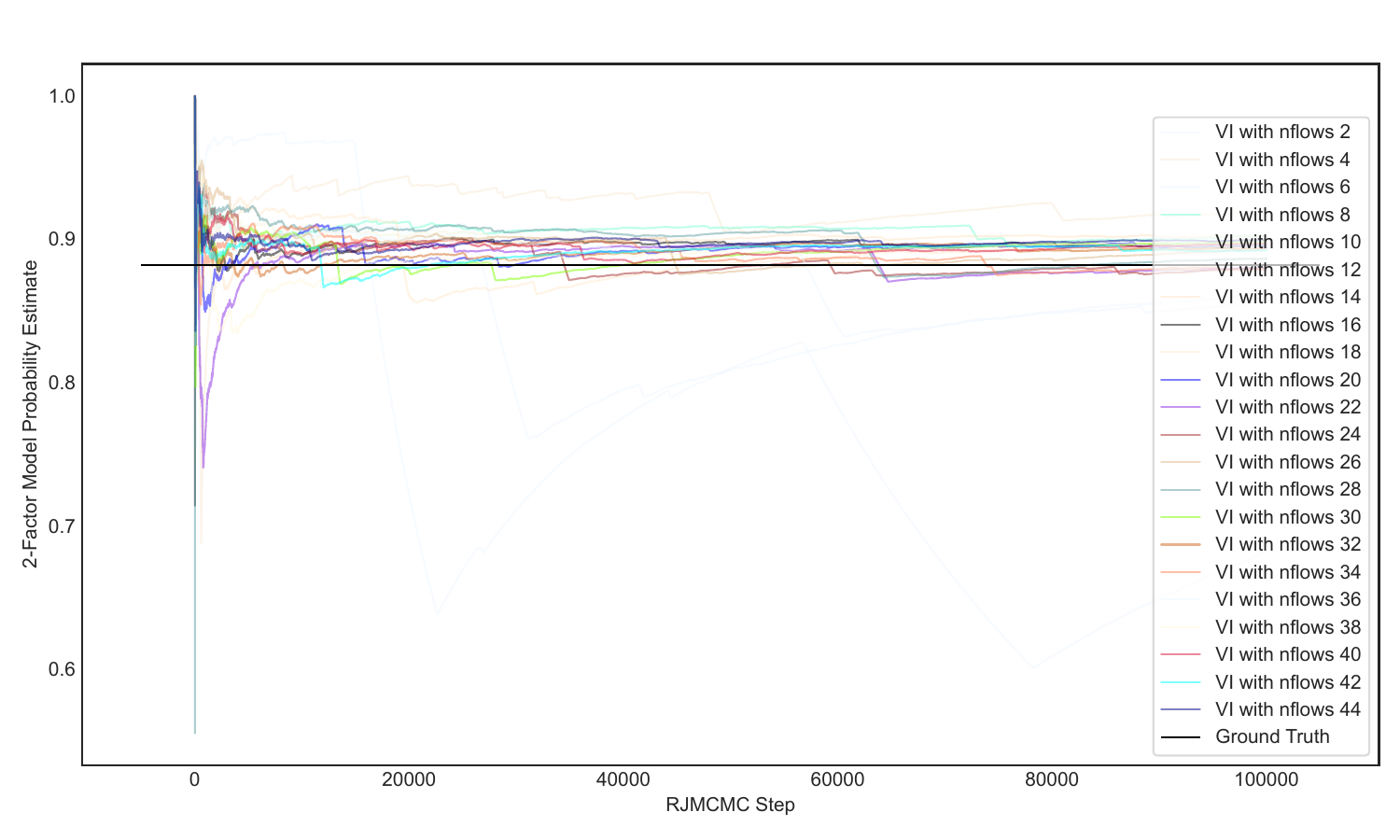}
  \caption{Flow-depth ablation for the factor-analysis example.  The panel documents the sensitivity of the VI-NFs TRJ running estimate to the number of RealNVP coupling layers.}
  \label{fig:FA-ablation-depth}
\end{figure}

\begin{figure}[!htbp]
  \centering
  \includegraphics[width=\linewidth]{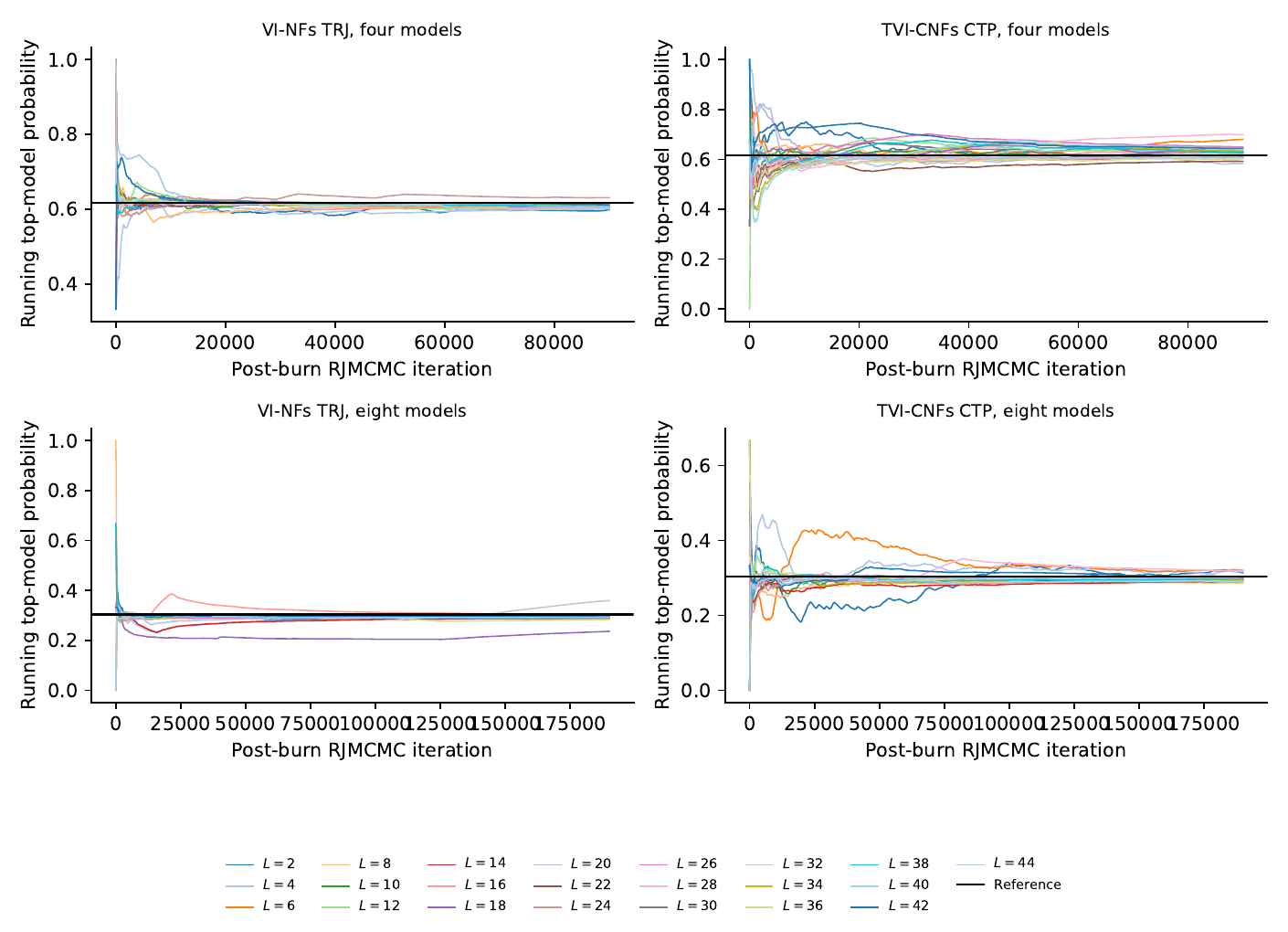}
  \caption{Flow-depth ablations for robust-regression variable selection.  The panels cover individual VI-NFs TRJ and TVI-CNFs CTP for the four-model and eight-model representations.  Each coloured curve is the mean running estimate over three RJMCMC chains after the first $10^4$ iterations are discarded; the black line is the corresponding reference probability.}
  \label{fig:VS-ablation-depth}
\end{figure}

\begin{figure}[!htbp]
  \centering
  \includegraphics[width=\linewidth]{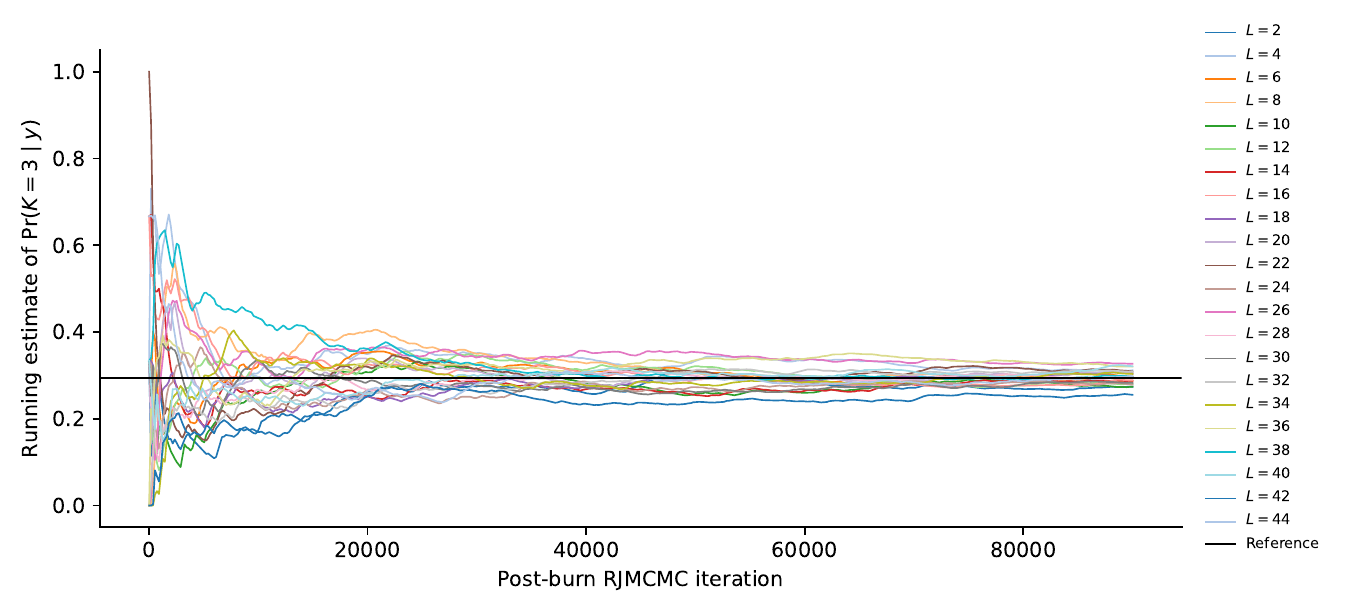}
  \caption{Flow-depth ablation for the coal-mining change-point example.  Each coloured curve is the mean running estimate of $\Pr(K=3\mid y)$ over three stored chains after a burn-in of $10^4$ iterations; the black line is the reference probability $0.294$.}
  \label{fig:CP-ablation-depth}
\end{figure}

\begin{figure}[!htbp]
  \centering
  \includegraphics[width=\linewidth]{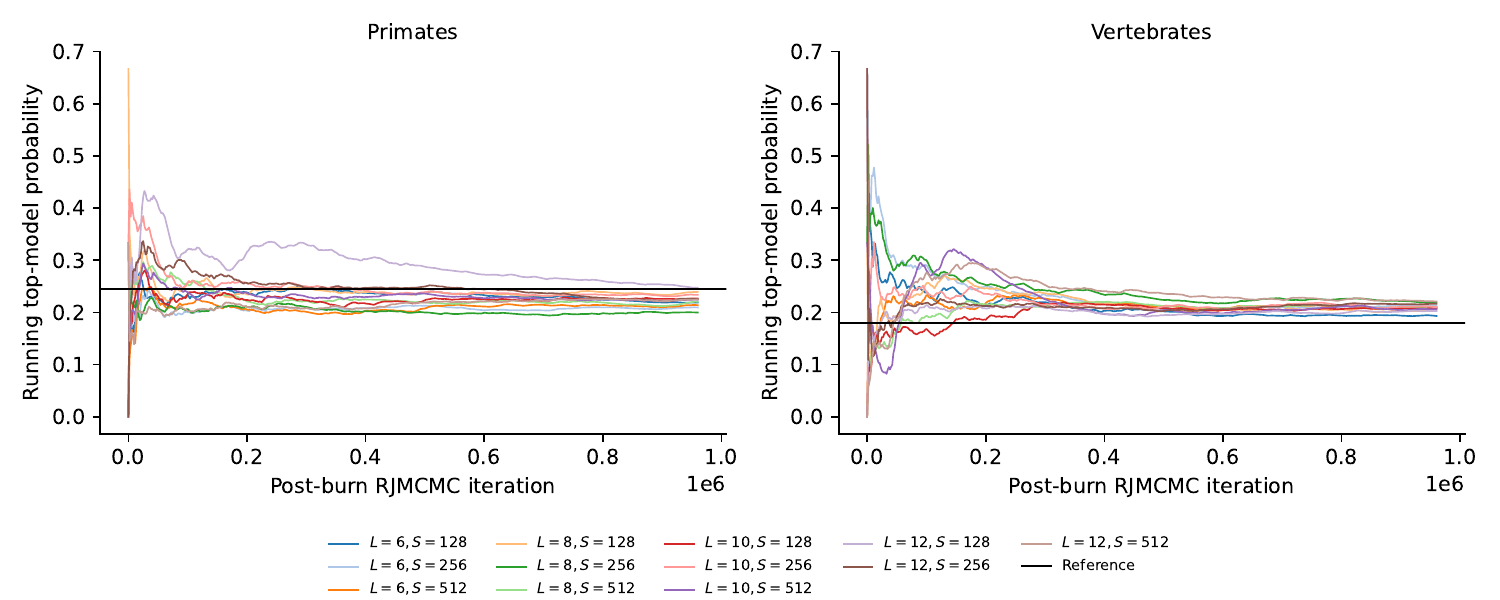}
  \caption{Flow-depth ablations for the phylogenetic substitution-model examples.  For each layer count and, the coloured curve is the mean running estimate of the displayed top-model probability over three stored chains after a 20\% burn-in; the black line is the corresponding reference probability.}
  \label{fig:PHYLO-ablation-depth}
\end{figure}

\FloatBarrier
\section*{Appendix D: Pilot Sampling and Map-Fitting Time}

All entries are seconds, reported as mean (SD) over the stored calibration/flow seeds.  Pilot sampling is a shared cost for the sample-trained methods.  The next two tables record map fitting after the corresponding pilot draws have been generated; the following table records direct reverse-KL optimization time.  Recorded phylogenetic sampling times are reported separately at the end of this appendix.

\begin{table}[!htbp]
\centering
\scriptsize
\caption{Factor-analysis shared pilot-draw generation in seconds, mean (SD) across three seeds.  The total is the paired SMC and NUTS wall time within each seed.}
\label{tab:FA-pilot-grid}
\begin{tabular}{lrrrr}
\toprule
Pilot component & $N=2000$ & $N=4000$ & $N=8000$ & $N=16000$ \\
\midrule
SMC, two-factor model & 38.3 (21.8) & 46.6 (1.6) & 100.3 (5.2) & 214.2 (2.2) \\
NUTS, three-factor model & 63.7 (15.2) & 88.3 (1.5) & 154.6 (1.2) & 288.8 (1.7) \\
Shared pilot total & 102.0 (36.7) & 134.9 (2.5) & 254.9 (5.3) & 503.0 (0.6) \\
\bottomrule
\end{tabular}
\end{table}

\begin{table}[!htbp]
\centering
\scriptsize
\caption{Shared SMC pilot-draw generation in seconds, mean (SD) across three seeds, for the variable-selection and change-point settings.  The labels VS/VSex and VSC/VSCex identify the stored ordinary and saturated representations, respectively.}
\label{tab:VSCP-pilot-grid}
\begin{tabular}{lrrrr}
\toprule
Example & $N=1000$ & $N=2000$ & $N=4000$ & $N=8000$ \\
\midrule
VS, four models & 5.7 (0.2) & 9.2 (0.9) & 17.9 (0.1) & 30.2 (0.2) \\
VSex, eight models & 15.7 (0.6) & 25.2 (2.0) & 47.8 (0.0) & 81.7 (0.3) \\
VSC, four-model saturated & 6.0 (0.3) & 9.2 (0.8) & 17.9 (0.1) & 30.3 (0.2) \\
VSCex, eight-model saturated & 16.4 (0.6) & 25.2 (2.1) & 48.0 (0.3) & 81.7 (0.4) \\
Change point & 90.4 (28.0) & 202.4 (16.2) & 399.8 (10.8) & 797.1 (23.6) \\
\bottomrule
\end{tabular}
\end{table}

\begin{table}[!htbp]
\centering
\scriptsize
\caption{Factor-analysis sample-trained proposal map-fitting time in seconds, mean (SD) across three seeds.  These are first-chain records after pilot draws are loaded and can include non-flow proposal setup; later chains reuse the fitted maps.}
\label{tab:FA-baseline-cold-setup-grid}
\begin{tabular}{lrrrr}
\toprule
Method & $N=2000$ & $N=4000$ & $N=8000$ & $N=16000$ \\
\midrule
Affine TRJ & 0.5 (0.0) & 0.8 (0.0) & 1.5 (0.0) & 2.8 (0.0) \\
RQMA-NF TRJ & 110.9 (18.2) & 110.0 (7.2) & 128.2 (12.5) & 111.1 (1.4) \\
\bottomrule
\end{tabular}
\end{table}

\begin{table}[!htbp]
\centering
\scriptsize
\caption{Sample-trained proposal map-fitting time in seconds, mean (SD) across three seeds, for the variable-selection and change-point baselines.  The first-chain record is shown after pilot draws are loaded; later chains reuse the fitted map.  \emph{CNF} denotes the sample-trained conditional rational-quadratic map in the stored implementation, rather than a target-trained VI map.}
\label{tab:VSCP-baseline-cold-setup-grid}
\resizebox{\linewidth}{!}{%
\begin{tabular}{llrrrr}
\toprule
Representation & Method & $N=1000$ & $N=2000$ & $N=4000$ & $N=8000$ \\
\midrule
VS & Affine TRJ & 1.8 (2.8) & 0.3 (0.0) & 0.4 (0.1) & 0.5 (0.1) \\
VS & RQMA-TRJ & 112.1 (2.9) & 110.7 (1.0) & 122.9 (20.2) & 111.5 (2.8) \\
VS & RQMA-CTP  & 28.7 (7.1) & 16.2 (5.4) & 28.7 (10.5) & 21.9 (2.6) \\
VSex & Affine TRJ & 0.5 (0.0) & 0.7 (0.0) & 1.0 (0.0) & 1.7 (0.0) \\
VSex & RQMA-TRJ & 227.3 (13.9) & 224.7 (3.3) & 230.0 (17.6) & 224.4 (5.6) \\
VSex & RQMA-CTP & 19.6 (4.0) & 27.9 (6.2) & 27.0 (6.7) & 23.6 (7.8) \\
VSC & RQMA-CTP & 28.7 (7.8) & 15.6 (4.6) & 23.6 (8.5) & 21.2 (2.0) \\
VSCex & RQMA-CTP & 17.0 (0.4) & 27.4 (6.7) & 25.2 (7.1) & 23.0 (6.6) \\
Change point & RQMA-CTP & 22.6 (2.6) & 18.0 (0.7) & 17.1 (3.6) & 21.1 (2.9) \\
\bottomrule
\end{tabular}%
}
\end{table}

\begin{table}[!htbp]
\centering
\scriptsize
\caption{Direct reverse-KL VI map-training time in seconds. For the factor-analysis and conditional proposals, each entry is the mean (SD) over the stored fits across all displayed $N$ settings; the two individual variable-selection entries are single stored map fits, subsequently reused across the three displayed RJMCMC chains.}
\label{tab:VI-map-training-grid}
\begin{tabular}{llr}
\toprule
Example & Proposed method & Time (s) \\
\midrule
Factor analysis & VI-NFs TRJ & 1369.7 (18.4) \\
VS, four models & VI-NFs TRJ & 2469.6 (14.6) \\ 
VS, four models & TVI-CNFs CTP & 2311.9 (52.3) \\
VSex, eight models & TVI-CNFs CTP & 2314.5 (61.0) \\
Change point & TVI-CNFs CTP & 2197.4 (115.7) \\
\bottomrule
\end{tabular}
\end{table}

\begin{table}[!htbp]
\centering
\scriptsize
\caption{Recorded RJMCMC sampling-phase time in seconds for the retained 1M-iteration phylogenetic chains, mean (SD) across three chains.}
\label{tab:PHYLO-sampling-time}
\begin{tabular}{llr}
\toprule
Dataset & Method & Time(s) \\
\midrule
Primates & Standard & 6256.8 (46.5) \\
Primates & TVI-CNFs CTP & 11771.5 (4882.1) \\
Vertebrates & Standard & 8852.0 (799.3) \\
Vertebrates & TVI-CNFs CTP & 9873.0 (1438.5) \\
\bottomrule
\end{tabular}
\end{table}